\documentclass{bmvc2k}
\usepackage{wrapfig}
\usepackage{amsfonts}
\usepackage{mdframed}
\usepackage{hyperref}

\title{Leveraging Modality Tags for Enhanced Cross-Modal Video Retrieval}

\addauthor{Adriano Fragomeni}{adriano.fragomeni@bristol.ac.uk}{1}
\addauthor{Dima Damen}{dima.damen@bristol.ac.uk}{1}
\addauthor{Michael Wray}{michael.wray@bristol.ac.uk}{1}

\addinstitution{
 School of Computer Science\\
 University of Bristol\\
 Bristol, UK
}

\runninghead{Fragomeni et al}{Leveraging Modality Tags for ... Video Retrieval}

\begin{document}

\maketitle

\begin{abstract} 
Video retrieval requires aligning visual content with corresponding natural language descriptions. In this paper, we introduce Modality Auxiliary Concepts for Video Retrieval (MAC-VR), a novel approach that leverages modality-specific tags -- automatically extracted from foundation models -- to enhance video retrieval.
We propose to align modalities in a latent space, along with learning and aligning auxiliary latent concepts derived from the features of a video and its corresponding caption.
We introduce these auxiliary concepts to improve the alignment of visual and textual latent concepts, allowing concepts to be distinguished from one another. 

We conduct extensive experiments on six diverse datasets: two different splits of MSR-VTT, DiDeMo, TGIF, Charades and YouCook2. The experimental results consistently demonstrate that modality-specific tags improve cross-modal alignment, outperforming current state-of-the-art methods across three datasets and performing comparably or better across others. Project Webpage: \url{https://adrianofragomeni.github.io/MAC-VR/}
\end{abstract}

\vspace{-15pt}
\section{Introduction}
\label{sec:intro}
\begin{wrapfigure}{r}{0.6\textwidth} 
    \centering
    \includegraphics[width=\linewidth]{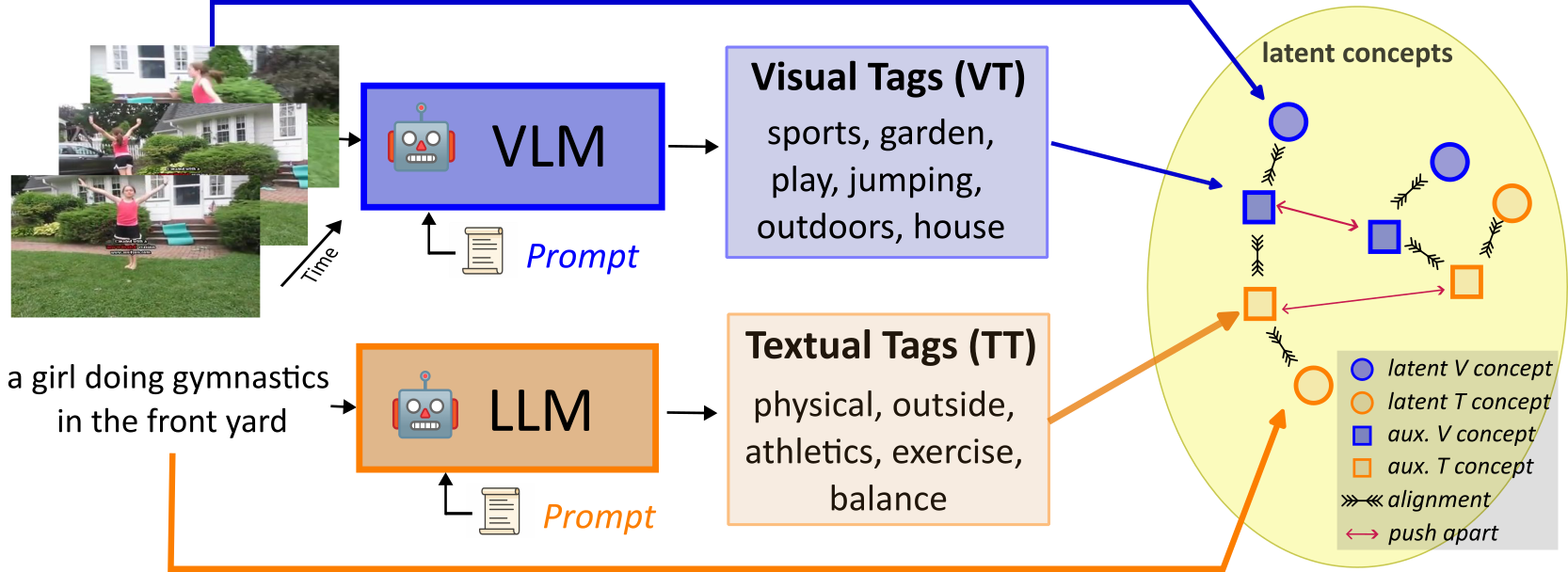}
        \vspace{-20pt}
    \caption{Tags are extracted from both \textcolor{blue}{videos} by VLM and \textcolor{orange}{texts} by LLM using prompts designed to generate tags for each modality. For example, \textcolor{blue}{sports} and \textcolor{orange}{physical} tags can help align this video and its caption.}
    \vspace{-10pt}
    \label{fig:intro}
\end{wrapfigure}

\vspace{-8pt}
The emergence of prominent video-sharing platforms, e.g., YouTube and TikTok, has supported uploading of millions of videos daily.
The demand for better video retrieval methods, which align textual queries with relevant video content, has subsequently increased. Most existing works use two main approaches. The first~\cite{fang2021clip2video,luo2022clip4clip,DBLP:conf/ijcai/JinLCHWYLC23} exclusively uses word and frame features.
However, the second~\cite{dzabraev2021mdmmt,gabeur2020multi,DBLP:conf/cvpr/WangZ021,DBLP:conf/wacv/GabeurN0AS22,DBLP:conf/iccv/Liu0QCDW21,DBLP:conf/iccv/CroitoruBLJZAL21} uses additional multi-modal information from videos -- such as audio, speech, and objects -- that is encoded and used for feature aggregation.  
In real-world scenarios, online videos often come with related textual information, such as tags -- keywords associated with a video that describe its content and make it easier to search/filter.
Few works~\cite{chen2023tagging,DBLP:conf/cvpr/WangGCY0SQS22,DBLP:conf/nips/WangLXZ00WY0HCB22} extract and exploit tags in video retrieval to better align the visual and textual modalities. These works only focus on extracting tags from the visual modality using pre-trained experts or a predefined visual token vocabulary. 
Inspired by these, we develop a novel method called MAC-VR that integrates multi-modal information by independently extracting tags \emph{for both videos and texts} without any manual annotation, utilising the extensive knowledge from pre-trained Vision-Language Models (VLM) and Large Language Models (LLM) as shown in Fig.~\ref{fig:intro}. 
The example in Fig.~\ref{fig:intro} shows the query ``a girl doing gymnastics in the front yard'', the extracted visual (VT) and textual (TT) tags include \textit{sports, physical, outdoors, and outside}, which can help align this video to its caption. 
We build on previous work~\cite{DBLP:conf/ijcai/JinLCHWYLC23}, due to its performance and efficiency, where visual and textual coarse features are split into compact latent factors which encode visual/textual concepts.
MAC-VR extends this work by introducing auxiliary modality-specific latent concepts, which are learnt from visual and textual tags using modality-specific foundation models.
These are aligned to the latent concepts extracted from videos and texts, through a novel Alignment Loss. 

Recently, many works~\cite{DBLP:conf/eccv/LiuXXCJ22,DBLP:conf/ijcai/JinLCHWYLC23,jin2023diffusionret,DBLP:conf/nips/JinHLWGSC022,DBLP:conf/iccv/IbrahimiSWGSO23,chen2023tagging} use different inference strategies to improve the final video retrieval performance such as Querybank Normalisation (QB)~\cite{DBLP:conf/cvpr/BogolinCJLA22} and Dual Softmax (DSL)~\cite{DBLP:journals/corr/abs-2109-04290}. In this paper, we analyse the impact of such strategies on our MAC-VR architecture to ensure fair comparison with state-of-the-art (SOTA) methods.
Our results show that auxiliary concepts of both modalities, in addition to the Alignment Loss, help boost the retrieval performance and better distinguish the latent concepts.

Our contribution is as follows:
    (i) We propose automatic extraction of modality-specific tags from foundational models to augment video/text.
    (ii) We use tags to learn auxiliary latent concepts in each modality to extract meaningful representations.
    (iii) We propose a new Alignment Loss to better align and distinguish these learnt latent concepts.
    (iv) We analyse the impact of different inference strategies, fairly comparing our proposal with SOTA methods.
    (v) We conduct experiments on six datasets: two splits of MSR-VTT, DiDeMo, TGIF, Charades, and YouCook2. Across all datasets, using auxiliary concepts improves performance over our baseline. A detailed ablation on MSR-VTT verifies our design.

\vspace{-16pt}
\section{Related works}
\vspace{-5pt}
\label{sec:related}
\textbf{Text-Video Retrieval.} Video retrieval aligns video and text through learned embeddings. Early methods~\cite{dzabraev2021mdmmt,gabeur2020multi,DBLP:conf/cvpr/WangZ021,DBLP:conf/wacv/GabeurN0AS22,DBLP:conf/iccv/Liu0QCDW21,DBLP:conf/iccv/CroitoruBLJZAL21,Fragomeni_2022_ACCV,DBLP:conf/iccv/ZolfaghariZGB21,DBLP:journals/corr/abs-2203-07086,DBLP:journals/pami/DongLXYYWW22} leveraged pre-trained features and multi-modal cues (e.g. audio, speech) to bridge the gap between videos and texts. Notably, MMT~\cite{gabeur2020multi} uses seven pre-trained experts in a brute-force fusion approach without explicit guidance.
Input modalities have also been masked~\cite{DBLP:conf/wacv/GabeurN0AS22} or enhanced by local temporal context~\cite{Fragomeni_2022_ACCV}. On the contrary, MAC-VR relies solely on video and text, without auxiliary modalities or context. Recent works have followed two main methodologies: involving extensive pre-training of models on large-scale video-text datasets,~\cite{DBLP:conf/cvpr/GeGLLSQL22,DBLP:conf/iccv/BainNVZ21} or transferring knowledge from image-based CLIP models~\cite{DBLP:journals/corr/abs-2103-00020} trained on extensive image-text pairs~\cite{DBLP:journals/corr/abs-2203-07086,fang2021clip2video,gorti2022xpool,luo2022clip4clip,jin2023diffusionret,Huang_2023_CVPR,Dong_2023_ICCV,guan2023pidro,tian2024holistic,jin2024mv,DBLP:conf/ijcai/JinLCHWYLC23,DBLP:conf/iclr/XueS0FSLL23,DBLP:conf/iccv/FangWLZS0SJW23}.
Some employ knowledge distillation from large teacher networks~\cite{Dong_2023_ICCV,tian2024holistic}, whereas MAC-VR bypasses distillation, training directly with auxiliary modality-specific tags. Similar to~\cite{DBLP:conf/ijcai/JinLCHWYLC23}, which learns latent queries and concepts, MAC-VR learns latent auxiliary concepts from these tags to enhance alignment between visual and textual features.\\
\textbf{Vision-Language and Large Language Models in Image and Video Retrieval.} The integration of Vision-Language Models (VLM)~\cite{li2023blip,liu2024visual,zhang2023video,cheng2024videollama,damonlpsg2023videollama} 
and Large Language Models (LLM)~\cite{touvron2023llama,touvron2023llama2,vicuna2023,DBLP:journals/corr/abs-2407-21783} has significantly advanced image~\cite{qu2024unified,DBLP:conf/nips/LevyBDL23,DBLP:conf/eacl/WangEDSB24,DBLP:conf/mir/ZhuHRK24,DBLP:conf/iccv/0003WZDHLWSM23} and video retrieval~\cite{DBLP:conf/cvpr/WuLFWO23,DBLP:conf/cvpr/0006MKG23,DBLP:conf/nips/WangLXZ00WY0HCB22,DBLP:conf/eccv/ShvetsovaKHRSK24,DBLP:journals/corr/abs-2401-00789,DBLP:journals/corr/abs-2401-06129,DBLP:conf/aaai/VenturaYSV24}, showcasing their  impressive understanding capabilities. 
Prior work enhances video retrieval by generating captions to enrich semantic context~\cite{DBLP:conf/cvpr/0006MKG23,DBLP:conf/cvpr/WuLFWO23}. In contrast, we leverage pre-trained  VLMs and LLMs to generate visual and textual tags that highlight key aspects of the video and its caption. Ablation studies confirm the superiority of tags over new captions extracted from the same models (see Sec.~\ref{sec:ablation}).\\
\textbf{Tags in Image and Video Understanding.} Tags have been used across various tasks, including Video Retrieval~\cite{DBLP:conf/cvpr/WangGCY0SQS22, chen2023tagging,DBLP:conf/nips/WangLXZ00WY0HCB22}, Video Moment Retrieval~\cite{DBLP:journals/tcsv/GaoX22, DBLP:conf/mm/WangWLY22}, Video Recognition~\cite{DBLP:conf/cvpr/WuWLWYO23, DBLP:journals/corr/abs-2304-02560}, Fashion Image Retrieval~\cite{DBLP:conf/mir/NakaKYG22, DBLP:conf/acl/Wang0LLCJHXG23, DBLP:conf/wacv/TianNB23, DBLP:conf/cvpr/ShimizuNG23, DBLP:conf/wacv/WahedZYL24} and Image Retrieval~\cite{DBLP:journals/corr/abs-2303-05657, DBLP:journals/corr/abs-2312-14149, DBLP:journals/tomccap/ChaudharyGGC20, DBLP:journals/tcsv/ZhuCCLZ21,DBLP:conf/eccv/ChiquierMV24}.
In~\cite{DBLP:conf/cvpr/WangGCY0SQS22,chen2023tagging,DBLP:conf/nips/WangLXZ00WY0HCB22}, tags are extracted using pre-trained experts from various modalities of videos, including object, person, scene, motion, and audio. 
Differently, MAC-VR leverages both visual and textual inputs, generating tags directly from videos and captions using VLMs and LLMs without relying on expert models or additional modalities. While~\cite{DBLP:conf/nips/WangLXZ00WY0HCB22} uses image-language models with a fixed visual token vocabulary to derive frame-level captions and attributes, MAC-VR avoids predefined vocabularies or caption generation, instead extracting tags end-to-end from raw inputs.

\vspace{-16pt}
\section{Modality Auxiliary Concepts for Video Retrieval}
\vspace{-8pt}
We first define text-to-video retrieval and inference strategies in Sec.~\ref{subsec:problem} before describing our tag extraction approach in Sec.~\ref{subsec:method:tags}.
Finally, in Sec.~\ref{subsec:method:architecture}, we introduce MAC-VR.
\vspace{-12pt}
\subsection{Cross-Modal Text-to-Video Retrieval}
\vspace{-5pt}
\label{subsec:problem}
Given a pair $(v_i,t_i)$, where $v_i$ represents a video and $t_i$ denotes its corresponding caption, the objective of Cross-Modal Text-to-Video Retrieval is to retrieve $v_{i}$ from a large gallery of videos, given the caption $t_{i}$ as query.
Typically, models use two 
projection functions: $f_v : v_i \longrightarrow \Omega \in \mathbb{R}^d$ and $f_t : t_i \longrightarrow \Omega \in \mathbb{R}^d$
to map the respective modalities into a shared d-dimensional latent embedding space, $\Omega$.
The aim is to align the representations in this space so that the representation of a video is close to that of its corresponding caption.
\begin{wrapfigure}{r}{0.55\textwidth} 
\centering
\vspace{-4pt}
\includegraphics[width=\linewidth]{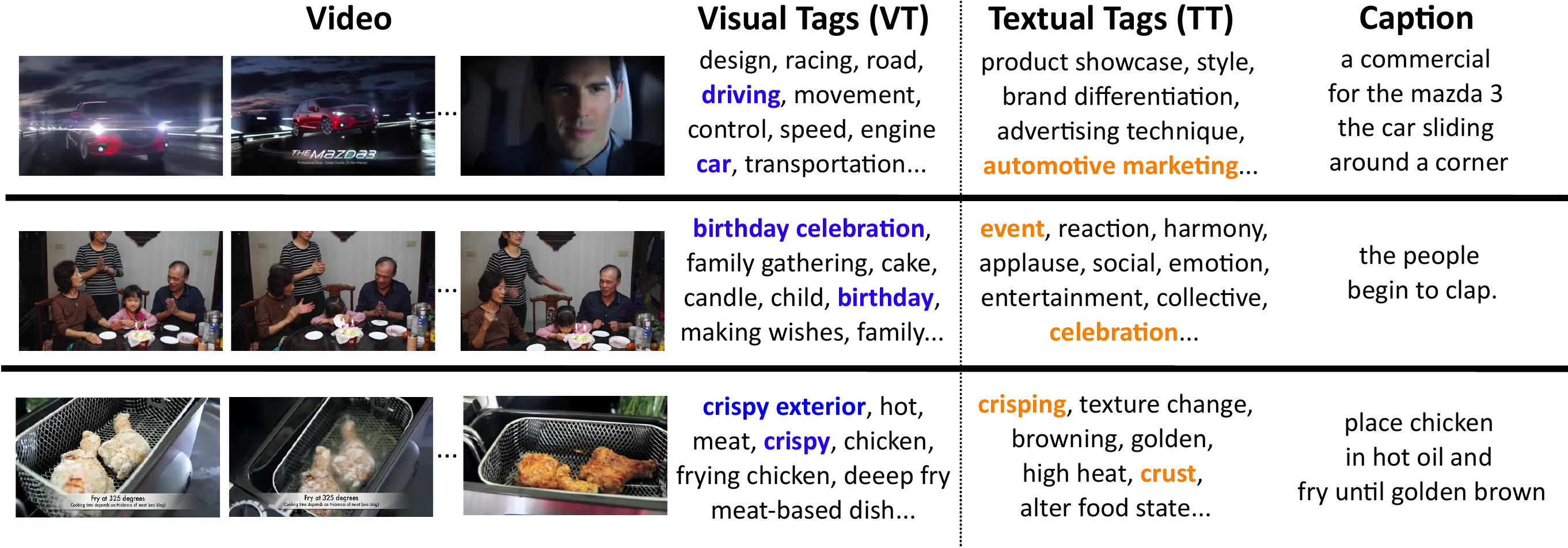}
    \vspace{-20pt}
    \caption{Examples of visual and textual tags (middle) for videos (left) and their captions (right). \textcolor{blue}{Blue} and \textcolor{orange}{orange} highlight semantic overlaps.}
    \vspace{-10pt}
    \label{fig:examples_tags}
\end{wrapfigure}
Following training, the standard inference strategy (IS) embeds a test video gallery and ranks these based on their distance from each query caption.
Recent approaches utilise additional inference strategies to improve performance: Querybank Normalisation (QB)~\cite{DBLP:conf/cvpr/BogolinCJLA22} and Dual Softmax (DSL)~\cite{DBLP:journals/corr/abs-2109-04290} are the most popular. We introduce these strategies in Sec.~\ref{appendix:inference_strategies} in supp. 

\vspace{-12pt}
\subsection{Tag Extraction}
\vspace{-5pt}
\label{subsec:method:tags}
We propose to extract tags automatically from either the video $v_i$, using a VLM, or the text in the caption $t_i$, using an LLM. These tags are word-level representations of common objects, actions, or general ideas present in the video or the caption. 
They can add additional useful information to retrieve the correct video given a text query as shown in Fig.~\ref{fig:examples_tags}. For instance, given the caption: \textit{a commercial for the Mazda 3 the car sliding around a corner}, the general tags estimated from this caption are: \textit{product showcase, advertising technique, automotive marketing}, which reflect the commercial. These words are terms that go beyond the exact caption but can help the model to understand the specific characteristics of this caption better. 

In contrast, leveraging the video modality to create tags enables us to both capture a broader array of visual elements that characterise the video content and also have a representation of the video in words, facilitating matching the video content to the captions within the text modality.
E.g., given the video associated with the previous caption, extracted visual tags include \textit{road, vehicle, car, transportation, engine}, reflecting important objects in the video, and \textit{racing, driving} reflecting the action in the video.
These tags directly correspond to pertinent visual components of the video. 
\begin{figure*}[t!]
    \centering
    \includegraphics[width=\linewidth]{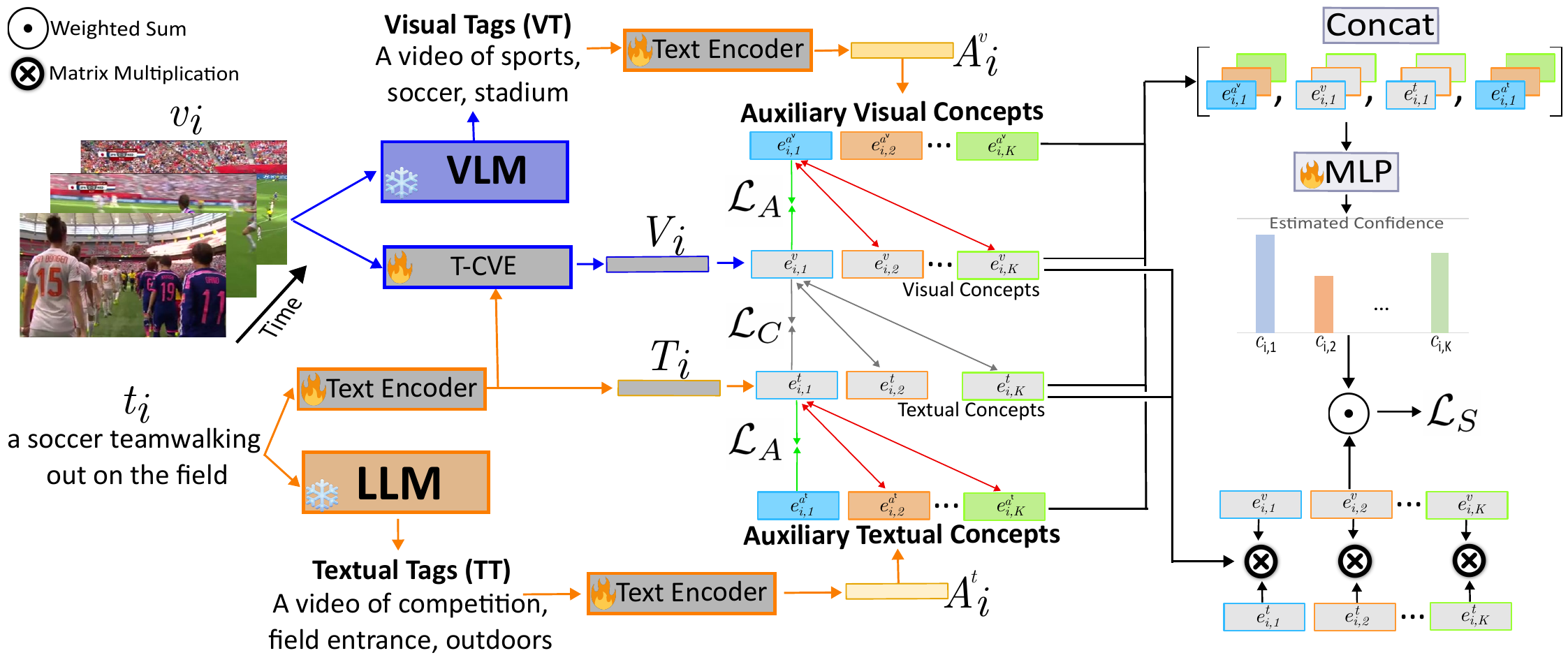}
    \vspace{-20pt}
    \caption{Architecture of MAC-VR: Given a video $v_i$ and its corresponding caption $t_i$, we generate auxiliary Visual Tags (VT) and Textual Tags (TT) using a VLM and an LLM, respectively. A shared text encoder projects the caption and the auxiliary tags $T_i$, $A_i^v$ and $A_i^t$ to a common space with the Text-Conditioned Video Encoder (T-CVE).
    Visual $e_{i,k}^v$ and textual $e_{i,k}^t$ concepts are aligned to each other by the contrastive loss $\mathcal{L}_C$ and are aligned to auxiliary visual $e^{a^v}_{i,k}$ and textual $e^{a^t}_{i,k}$ concepts by our Alignment Loss $\mathcal{L}_A$. An MLP then estimates confidence scores for each concept, to compute a weighted sum for the similarity function that is used in our  Cross-Modal Loss $\mathcal{L}_S$.}
    \label{fig:model}
\vspace{-12pt}
\end{figure*}
To extract visual and textual tags, we use a custom prompt (see Sec.~\ref{appendix:tag_extract} in supp.) to query the most general tags for the input video $v_i$ and caption $t_i$. We extract tags individually in an offline manner from both modalities so they can be used for training and inference.
We call these tags as modality-specific tags because they are extracted from a single modality. 
\vspace{-12pt}
\subsection{Architecture}
\vspace{-5pt}
\label{subsec:method:architecture}

We start from a standard Text-Conditioned Video Encoder (T-CVE) before incorporating our proposed tags into each modalities' latent concepts. These concepts are aligned and pooled to find the similarity between a video and a caption. Our architecture is shown in Fig.~\ref{fig:model}.
\vspace{-12pt}
\subsubsection{Text-Conditioned Video Encoder (T-CVE)}
\vspace{-5pt}
Given a caption $t_i$, we extract its text representation ${T_i \in \mathbb{R}^d}$.
For the video representation, we first sample $N_v$ frames from a video $v_i$, then encode them and aggregate the embedding of all frames to obtain the frame representation $F_j$ with $j \in \{1,..., N_v\}$.
Since captions often describe specific moments, as shown in previous works~\cite{bain2022clip,DBLP:conf/ijcai/JinLCHWYLC23,gorti2022xpool}, matching only the relevant frames improves semantic precision and reduces noise. To achieve this, we aggregate the frame representations conditioned on the text. Firstly, we calculate the inner product between the text and the frame representation $F_j$ with $j \in \{1,..., N_v\}$:\\
\begin{equation}
a_{i,j}=\frac{exp((T_i)^{\top} F_j/\tau_a)}{\sum^{N_v}_{k=1}(exp((T_i)^{\top} F_k/\tau_a))}
    \label{eqn:text_condition}
\end{equation}
where $\tau_a$ is a hyper-parameter that allows control of the textual conditioning. Then, we get the text-conditioned video representation $V_i \in R^d$ defined as $V_i=\sum\limits^{N_v}_{k=1}a_{i,k}F_k$.
\vspace{-12pt}
\subsubsection{Latent Concepts} 
\vspace{-5pt}
To utilise the visual and textual tags extracted from foundational models in Sec~\ref{subsec:method:tags},  we first randomly pick $N$ visual and textual tags (per modality) during training and order them into two distinct comma-separated sentences that start with \textit{``A video of''}. We extract visual $A_i^v$ and textual $A_i^t$ coarse tag features by using the same text encoder used for the caption. Therefore, given a video/caption pair $(v_i,t_i)$ we get a quadruple $(V_i, T_i, A_i^v, A_i^t)$. Inspired by~\cite{DBLP:conf/ijcai/JinLCHWYLC23}, we disentangle each element of the quadruple into $K$ independent, equal-sized latent concepts. For example, when disentangling $V_i$, we get $K$ independent latent concepts, i.e. $E_i^v=[e^v_{i,1},..., e^v_{i,K}]$. Each latent concept $e^v_{i,k} \in R^{d/K}$ represents a distinct concept and the independence of these factors ensures that each concept is uncorrelated to the other $K\text{-}1$ latent concepts, and is thus defined by independently projecting the video representation as $e^v_{i,k}=W^v_kV_i$,
where $W^v_k$ is a trainable parameter. 
Similarly, $E_i^t$, $E_i^{a^v}$ and $E_i^{a^t}$  represent the latent concepts of the text $T_i$, the visual $A_i^v$ and the textual $A_i^t$ tag representations, which are calculated in the same way. 
We name the $K$ latent concepts of the tags representation as the \textbf{auxiliary visual concepts} $E_i^{a^v}$ and \textbf{auxiliary textual concepts} $E_i^{a^t}$ respectively.
We now have four disentangled representations for visual $E_i^v$, textual $E_i^t$, auxiliary visual
$E_i^{a^v}$, and auxiliary textual $E_i^{a^t}$ concepts. 
Until now, these subspaces have been disentangled independently. 
We next describe how to align these latent concepts.
\vspace{-12pt}
\subsubsection{Alignment of Disentangled Latent Concepts}
\vspace{-5pt}
\label{sec:alignment_loss}

By default, approaches such as~\cite{DBLP:conf/ijcai/JinLCHWYLC23} directly align latent representations of videos and captions through a contrastive loss.
Here, we extend this by applying the contrastive loss to align modality-specific concepts to the corresponding auxiliary concepts. In this way, modality-specific concepts assist MAC-VR in better aligning the video and text modalities.
Specifically, we consider the visual concepts $E_i^v$ and the auxiliary visual concepts $E_i^{a^v}$.
For each disentangled concept pair $(e_{i,k}^v, e_{i,k}^{a^v})$ we minimise the distance between this pair then maximise the distance to other disentangled concepts, i.e. $(e_{i,k}^v, e_{i,l}^{a^v}):\, l \ne k$ in a contrastive fashion to align modality concepts. 
Recall that these latent concepts are learnt, and thus through this alignment, we aim to learn a representation of the video that matches the latent representations of the tags extracted from the VLM.
Similarly, we align the auxiliary textual concepts $E_i^{a^t}$ to the latent concepts $E_i^t$ extracted directly from the caption.
We combine both modalities' alignment of latent concepts to auxiliary latent concepts and refer to this as the Alignment Loss $\mathcal{L}_A$ which aligns $E_i^v$ with $E_i^{a^v}$ and $E_i^t$ with $E_i^{a^t}$.
\vspace{-12pt}
\subsubsection{Weighted Similarity and Training Loss}
\vspace{-5pt}

Only a subset of visual concepts are usually described in the corresponding text.
Therefore, we cannot directly leverage correlations between their latent concepts, so we use adaptive pooling to define weights for the visual and textual concepts and reduce their impact on the final similarity calculation. To do this, we design an adaptive module to estimate the confidence of each cross-modal concept matching. 
For each concept $k$, we concatenate the modality and auxiliary modality concepts $[e^v_{i,k},e^t_{i,k},e^{a^v}_{i,k},e^{a^t}_{i,k}]$ and use them to calculate the confidence of each cross-modal concept matching $c_{i,k}=MLP([e^v_{i,k},e^t_{i,k},e^{a^v}_{i,k},e^{a^t}_{i,k}])$.
If the confidence $c_{i,k}$ is low, the corresponding latent concept of the $k^{th}$ subspace is matched with a low score. 
Given this confidence, we aggregate visual and textual latent concept pairs to calculate the overall similarity of the video and text $S(v_i, t_i)$, through adaptive pooling as:
\begin{equation}
S(v_i, t_i)=\sum^K_{k=1}c_{i,k}\cdot\frac{(e^t_{i,k})^{\top}e^v_{i,k}}{|\|e^t_{i,k}\|\|e^v_{i,k}\|}
    \label{eqn:confidence}
\end{equation}
Following common approaches, we use the InfoNCE loss~\cite{DBLP:journals/jmlr/GutmannH12,DBLP:journals/corr/JozefowiczVSSW16} as our Cross-Modal Loss ($\mathcal{L}_{S}$) to optimise the cross-modal similarity $S(v_i, t_i)$; the contrastive loss $\mathcal{L}_C$ to align the latent modality concepts (as introduced in~\cite{DBLP:conf/ijcai/JinLCHWYLC23}); and our proposed Alignment Loss $\mathcal{L}_{A}$ to align the modality with the auxiliary modality concepts. We define our total loss as $\mathcal{L}=\mathcal{L}_S(S(v_i, t_i))+ \alpha_C\mathcal{L}_{C}(E_i^v, E_i^t) +\alpha_A \mathcal{L}_{A}(E_i^v, E_i^t, E^{a^v}_i, E^{a^t}_i)$,
where $\alpha_C$ and $\alpha_A$ are hyperparameters to weigh the losses.
During inference, the weighted similarity $S(v_i, t_i)$ as in Eq.~\ref{eqn:confidence} is used, given the query caption and every video in the gallery.
\vspace{-16pt}
\section{Experiments}
\vspace{-5pt}
\subsection{Datasets, Implementation Details and Metrics}
\vspace{-5pt}
\label{sec:data_details_metrics}
\noindent\textbf{Datasets:} We perform experiments on five video datasets: MSR-VTT~\cite{xu2016msr}, DiDeMo~\cite{DBLP:journals/corr/abs-1708-01641}, TGIF~\cite{DBLP:conf/cvpr/LiSCTGJL16}, Charades~\cite{DBLP:conf/eccv/SigurdssonVWFLG16} and YouCook2~\cite{DBLP:conf/aaai/ZhouXC18}, more details in Sec.~\ref{appendix:datasets} of the supp.\\
\noindent\textbf{Architecture Details:} We build our model using DiCoSA~\cite{DBLP:conf/ijcai/JinLCHWYLC23} as our baseline, employing CLIP’s VIT-B/32~\cite{radford2021learning} as the image encoder and CLIP’s transformer base as the text encoder to encode captions and visual/textual tags. Unless specified, we set $\tau_a=3$ in Eq.~\ref{eqn:text_condition}, $K=8$, and weight parameters $\alpha_C=\alpha_A=1$. Our MLP consists of two 256-length linear layers with a ReLU activation, and we employ a 4-layer transformer to aggregate frame embeddings.\\ \textbf{Tag Extraction:} We use the fine-tuned VideoLLaMA2~\cite{cheng2024videollama} for visual tags and Llama3.1~\cite{DBLP:journals/corr/abs-2407-21783} for textual tags. VideoLLaMA2 runs on $8$ sparsely sampled frames with temperature values $\tau \in {0.7, 0.8, 0.9, 1.0}$ -- the same values are used for Llama3.1 -- to yield diverse tags. During training, we randomly select $N$ tags, while during inference we use fixed $M$ tags for deterministic results;  we set $N=M=12$ for training and evaluating MAC-VR, except when using DSL, where $N=6$ and $M=8$. We extract modality-specific tags from all videos and captions in the datasets above. On average, we have $27$--$29$ visual and textual tags per video/caption across all the datasets and splits. More tag extraction details in supp.\\ 
\noindent\textbf{Train/Val Details:} Following~\cite{DBLP:conf/ijcai/JinLCHWYLC23}, we optimise the model using an Adam optimiser with linear warm-up, setting initial learning rates of 1e-7 for the text and video encoders and 1e-3 for other modules. We use $N_v=12$ frames and $BS=10$ for all datasets except DiDeMo, where $N_v=50$ for video-paragraph retrieval and $BS=64$ due to GPU memory.\\
\noindent\textbf{Metrics:}~We present the retrieval performance for text-to-video retrieval task using standard metrics: Recall at $L = 1, 5$ ($R@L$) and mean rank ($MeanR$). We report the complete tables including $R@10$ and median rank $MR$ in Sec.~\ref{appendix:tables} in the supp. We use the sum of cross-modal $R@L$ with $L=1,5,10$ as $RSum$ to rank methods. 

\vspace{-12pt}
\subsection{Results}
\label{sec:results}
\vspace{-5pt}
First, we present the comparison of MAC-VR against the baseline DiCoSA~\cite{DBLP:conf/ijcai/JinLCHWYLC23}.
\begin{figure}[h!]
\centering
\includegraphics[width=\linewidth]{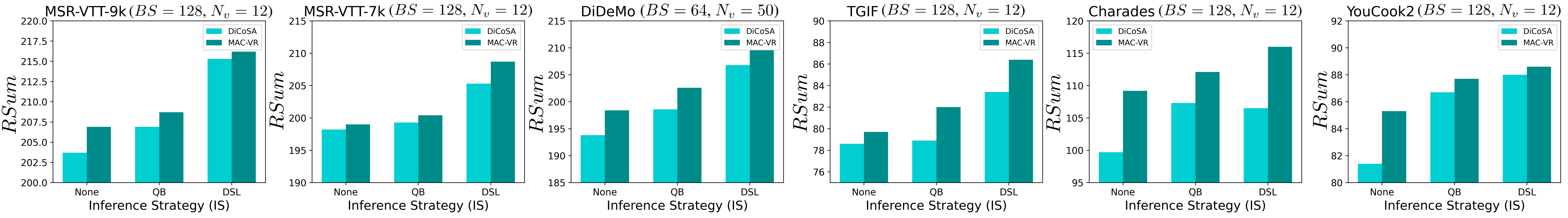}
\vspace*{-24pt}
    \caption{Comparison of $RSum$ with baseline. $BS$: Batch Size. $N_v$: Number of Frames.}
    \label{fig:comparison_baseline}
    \vspace*{-8pt}
\end{figure}
\begin{table*}[ht!]
\centering
\begin{minipage}{0.48\linewidth}
\centering
\resizebox{\linewidth}{!}{%
\begin{tabular}{lccc}
\hline
Dataset & DiCoSA (s/sample) & MAC-VR (s/sample) & $\Delta$ Time (s/sample) \\
\hline
Charades & 0.023 $\pm$ 0.001 & 0.057 $\pm$ 0.002 & +0.034 \\
DiDeMo   & 0.143 $\pm$ 0.003 & 0.179 $\pm$ 0.004 & +0.036 \\
MSRVTT   & 0.018 $\pm$ 0.001 & 0.065 $\pm$ 0.006 & +0.047 \\
TGIF     & 0.055 $\pm$ 0.001 & 0.077 $\pm$ 0.002 & +0.022 \\
YC2      & 0.020 $\pm$ 0.003 & 0.061 $\pm$ 0.001 & +0.042 \\
\hline
\end{tabular}}
\caption{Inference time comparison between DiCoSA and MAC-VR per sample.}
\label{tab:inference_time}
\end{minipage}
\hfill
\begin{minipage}{0.48\linewidth}
\centering
\resizebox{0.8\linewidth}{!}{%
\begin{tabular}{lcc}
\hline
Dataset & VT (s/sample) & TT (s/sample) \\
\hline
Charades & 0.918 $\pm$ 0.103 & 0.964 $\pm$ 0.147 \\
DiDeMo   & 0.927 $\pm$ 0.091 & 0.981 $\pm$ 0.021 \\
MSRVTT   & 0.971 $\pm$ 0.319 & 0.941 $\pm$ 0.186 \\
TGIF     & 0.555 $\pm$ 0.100 & 0.926 $\pm$ 0.116 \\
YC2      & 1.043 $\pm$ 0.131 & 1.016 $\pm$ 0.141 \\
\hline
\end{tabular}}
\caption{Visual and textual tag extraction times per sample for a single $\tau$.}
\label{tab:extraction_time}
\end{minipage}
    \vspace*{-14pt}
\end{table*}
Then, we compare MAC-VR with SOTA methods, particularly highlighting the impact of inference strategies on fairness of comparison.\\
\textbf{Comparison with Baseline.} In Fig.~\ref{fig:comparison_baseline}, we compare MAC-VR with our Baseline DiCoSA trained using the same training parameters, i.e. batch size, $BS$; same number of frames, $N_v$; and when using all inference strategies (IS). 
In all the datasets we improve $\Delta RSum$ compared to our baseline DiCoSA across all IS -- $+0.9$ for MSR-VTT-9k, $+3.4$ for MSR-VTT-7k, $+2.8$ for DiDeMo, $+9.5$ for Charades, $+3.0$ for TGIF and $+0.6$ for YouCook2 when using DSL. Tab.\ref{tab:inference_time} presents the per-sample inference time comparison between DiCoSA\cite{DBLP:conf/ijcai/JinLCHWYLC23} and MAC-VR across all datasets, indicating that while MAC-VR increases its inference time slightly, it remains comparable to the original DiCoSA. Furthermore, Tab.~\ref{tab:extraction_time} reports the extraction time of visual and textual tags for a single $\tau$ across datasets, which should be considered as an additional component of the total inference time.\\
\textbf{Comparison with SOTA.} In Tab.~\ref{tab:comparison_sota}, we provide a comprehensive comparison of MAC-VR against all SOTA works.
To ensure fairness, we divide methods by inference strategy (IS) and evaluate three settings: MAC-VR without IS, with QB, and with DSL.
We exclude methods that use additional input modalities or stronger pre-training. For instance, in \cite{DBLP:conf/iccv/IbrahimiSWGSO23,DBLP:conf/wacv/GabeurN0AS22,DBLP:conf/nips/ChenLWZSZL23,DBLP:conf/eccv/LinLBB22}, models also train on audio, so they are not directly comparable to MAC-VR, which only uses video and text. 
Moreover, VAST~\cite{DBLP:conf/nips/ChenLWZSZL23} includes subtitles (when available) in training as 
additional modality and is pre-trained on a multi-modal caption dataset.
Similarly, CLIP-ViP~\cite{DBLP:conf/iclr/XueS0FSLL23} uses extra pre-training datasets to improve retrieval performance further.
We also omit T-MASS~\cite{wang2024text} as the official code has been retracted due to data leakage.
Overall, MAC-VR achieves best overall performance when combined with DSL across all the datasets.
Results are consistent using QB for MSR-VTT (both splits) and without inference strategies on TGIF, Charades and YouCook2. On MSR-VTT (both splits), results are comparable to SOTA without IS. Across 5 of the 6 datasets, MAC-VR shows the best performance for any inference strategy. However, on DiDeMo, its performance is lower than SOTA methods for all inference strategies, even though the DSL approach still gives good results. This difference may be influenced by different training parameters (i.e. $BS$ and $N_v$) used in other SOTA~\cite{DBLP:conf/nips/JinHLWGSC022,guan2023pidro,jin2023video,jin2023diffusionret,DBLP:conf/cvpr/WuLFWO23,DBLP:conf/iccv/WangSCBB23,DBLP:conf/nips/LiSGZS23,chen2023tagging}. To showcase this, we re-ran two SOTA methods~\cite{DBLP:conf/nips/JinHLWGSC022,DBLP:conf/iccv/WangSCBB23} using our training settings, and both performed worse than in their original experiments, i.e. $\text{-}6.4$ for~\cite{DBLP:conf/nips/JinHLWGSC022} and $\text{-}15.3$ for~\cite{DBLP:conf/iccv/WangSCBB23}, and worse than our MAC-VR, i.e. $\text{-}0.6$ for~\cite{DBLP:conf/nips/JinHLWGSC022} and $\text{-}8.0$ for~\cite{DBLP:conf/iccv/WangSCBB23}.
\begin{table*}[ht!]
\centering
\resizebox{\linewidth}{!}{
\begin{tabular}{lcc|cccc|cccc|cccc}

&&&\multicolumn{4}{c|}{\textbf{MSR-VTT-9k}}&\multicolumn{4}{c|}{\textbf{MSR-VTT-7k}}&\multicolumn{4}{c}{\textbf{DiDeMo}}\\\cline{4-15}

\multicolumn{1}{l}{Method}&\multicolumn{1}{c}{IS}&\multicolumn{1}{c|}{Venue}
&\multicolumn{1}{c}{$R@1\uparrow$} 
&\multicolumn{1}{c}{$R@5\uparrow$}
&\multicolumn{1}{c}{$MeanR\downarrow$}
&\multicolumn{1}{c}{$RSum\uparrow$}
&\multicolumn{1}{|c}{$R@1\uparrow$} 
&\multicolumn{1}{c}{$R@5\uparrow$}
&\multicolumn{1}{c}{$MeanR\downarrow$}
&\multicolumn{1}{c}{$RSum\uparrow$}
&\multicolumn{1}{|c}{$R@1\uparrow$} 
&\multicolumn{1}{c}{$R@5\uparrow$}
&\multicolumn{1}{c}{$MeanR\downarrow$}
&\multicolumn{1}{c}{$RSum\uparrow$}\\\hline
CLIP4Clip~\cite{DBLP:journals/ijon/LuoJZCLDL22}&-&Neurocomp.'22&44.5&71.4&15.3&197.5&42.1&71.9&16.2&195.4&43.4&70.2&17.5&194.2\\
CenterCLIP~\cite{DBLP:conf/sigir/ZhaoZWY22}&-&SIGIR'22 & 44.2 & 71.6 & 15.1 &197.9&43.7&71.3&16.9&195.8& -&-&-&-\\
X-Pool~\cite{gorti2022xpool} &-& CVPR'22 & 46.9 & 72.8 & 14.3&201.9 &43.9&72.5&\textbf{14.6}&198.7&-&- & - & - \\
TS2-Net~\cite{DBLP:conf/eccv/LiuXXCJ22} &-&ECCV'22 & 47.0 & 74.5 & 13.0&205.3&-&-&-&-& 41.8 & 71.6 & 14.8&195.4\\
EMCL-Net~\cite{DBLP:conf/nips/JinHLWGSC022} &-& NeurIPS'22 & 46.8 & 73.1 & - &203.0&-&-&-&-&46.8&74.3 & 12.3 & 204.2 \\
EMCL-Net$^*$~\cite{DBLP:conf/nips/JinHLWGSC022} &-& NeurIPS'22 &-&-&-&-&45.2&71.4&15.0&197.8&44.1&71.7&15.1&197.8\\
VoP~\cite{Huang_2023_CVPR}&-&CVPR'23 & 44.6 & 69.9 & 16.3&194.8&42.7&68.2&15.9&190.2& 46.4 & 71.9 & 13.6&199.8\\
MuMUR~\cite{DBLP:journals/ir/MadasuASRTBL23}&-&Inf. Retr. J.'23&46.4&72.6&13.9&201.2&44.8&\textbf{72.0}&-&\textbf{199.3}&44.4&74.3&-&201.8\\
PiDRo~\cite{guan2023pidro} &-&ICCV'23& 48.2 & 74.9 & 12.6&206.4&-&-&-&-& 48.6 & 75.9 & 11.8&208.9\\
HBI~\cite{jin2023video} &-&CVPR'23 & 48.6 & 74.6 & \textbf{12.0}&206.6&-&-&-&-& 46.9 & 74.9 & 12.1 &204.5\\
DiffusionRet~\cite{jin2023diffusionret} &-& ICCV'23 & 49.0 & 75.2 & 12.1&206.9&-&-&-&-& 46.7 & 74.7 & 14.3&204.1  \\
Prompt Switch~\cite{deng2023prompt} &-& ICCV'23& 47.8 & 73.9 & 14.4&203.9&-&-&-&-& - & - & - & - \\
Cap4Video~\cite{DBLP:conf/cvpr/WuLFWO23} &-& CVPR'23& 49.3 & 74.3 & \textbf{12.0}&\textbf{207.4}&-&-&-&-& \textbf{52.0} & \textbf{79.4} & \textbf{10.5} &\textbf{218.9}\\
UCoFiA~\cite{DBLP:conf/iccv/WangSCBB23}&-& ICCV'23& \textbf{49.4} & 72.1 & 12.9&205.0&-&-&-&-& 46.5 & 74.8 & 13.4&205.7\\
UCoFiA$^*$~\cite{DBLP:conf/iccv/WangSCBB23}&-& ICCV'23&-&-&-&-&45.2&69.6&15.7&194.2&42.1&69.2&16.3&190.4\\
PAU~\cite{DBLP:conf/nips/LiSGZS23} &-& NeurIPS'23& 48.5 & 72.7 & 14.0 &203.7&-&-&-&-& 48.6 & 76.0 & 12.9&209.1\\
TABLE~\cite{chen2023tagging}&-& AAAI'23 & 47.1& 74.3 & 13.4&204.3&-&-&-&-& 47.9 & 74.0 & 14.3&204.0\\
UATVR~\cite{DBLP:conf/iccv/FangWLZS0SJW23}&-& ICCV'23 & 47.5& 73.9 & 12.3&204.9&-&-&-&-& 43.1 & 71.8 & 15.1&197.2\\
TeachCLIP~\cite{tian2024holistic} &-& CVPR'24& 46.8 & 74.3 & - & - &-&-&-&-& 43.7 & 71.2 & - & -\\
MV-Adapter~\cite{jin2024mv} &-& CVPR'24& 46.2 & 73.2 & - &202.1&-&-&-&-& 44.3 & 72.1 & - &196.9\\
BiC-Net~\cite{DBLP:journals/tomccap/HanZSXCC24}&-&TOMM'24&39.4&\textbf{75.5}&-&201.6&32.8&68.2&-&183.4&-&-&-&-\\
RAP~\cite{DBLP:conf/acl/CaoTHJ0L0LY024}&-&ACL'24&44.8&71.4&14.4&197.7&-&-&-&-&42.6&70.4&18.0&192.6\\
\textbf{MAC-VR (ours)} &-& - & 48.8 & 74.4 & 12.3 &206.9&\textbf{45.3}&71.9&15.0&199.0&43.4&72.7&16.9&198.4\\\hline

QB-Norm~\cite{DBLP:conf/cvpr/BogolinCJLA22} &QB& CVPR'22& 47.2 & 73.0 & -&203.2&-&-&-&-& 43.3 & 71.4 & -&195.5\\
DiCoSA~\cite{DBLP:conf/ijcai/JinLCHWYLC23} &QB& IJCAI'23& 47.5 & 74.7 & 13.2&206.0&-&-&-&-& 45.7 & 74.6 & \textbf{11.7}&203.8 \\
DiffusionRet~\cite{jin2023diffusionret} &QB& ICCV'23 & 48.9 & 75.2 & \textbf{12.1}&207.2&-&-&-&-& \textbf{48.9} & \textbf{75.5}& 14.1  &\textbf{207.7}\\
\textbf{MAC-VR (ours)}&QB& - & \textbf{49.3}& \textbf{75.9} & 12.3 &\textbf{208.7}&\textbf{45.6}&\textbf{72.7}&\textbf{16.0}&\textbf{200.4}&45.5&74.8&16.2&202.6\\\hline

EMCL-Net~\cite{DBLP:conf/nips/JinHLWGSC022} &DSL& NeurIPS'22 & 51.6 & 78.1 & - &215.0&-&-&-&-&-&-& - & -\\
EMCL-Net$^*$~\cite{DBLP:conf/nips/JinHLWGSC022} &DSL& NeurIPS'22 &-&-&-&-&-&-&-&-&47.6&73.5&\textbf{11.9}&203.9\\
TS2-Net~\cite{DBLP:conf/eccv/LiuXXCJ22} &DSL& ECCV'22 & 51.1 & 76.9 & 11.7 &213.6&-&-&-&-& 47.4 & 74.1 & 12.9&203.9\\
TABLE~\cite{chen2023tagging}&DSL& AAAI'23 & 52.3 & \textbf{78.4} & 11.4&215.9&-&-&-&-& 49.1 & \textbf{75.6} & 14.8 &207.6\\
UATVR~\cite{DBLP:conf/iccv/FangWLZS0SJW23}&DSL& ICCV'23 & 49.8& 76.1 & 12.3&211.4&-&-&-&-&-&-& - & - \\
RAP~\cite{DBLP:conf/acl/CaoTHJ0L0LY024}&DSL& ACL'24&-&-&-&-&-&-&-&-&47.1&74.1&13.9&203.6\\
\textbf{MAC-VR (ours)}&DSL& - &\textbf{53.2}  & 77.7 & \textbf{10.0}&\textbf{216.2}&\textbf{49.8}&\textbf{74.6}&\textbf{12.7}&\textbf{208.7}&\textbf{50.2}&75.2&15.1&\textbf{209.6}\\\hline\hline
\end{tabular}}

\vspace*{2pt}

\resizebox{\linewidth}{!}{
\begin{tabular}{lcc|cccc|cccc|cccc}

&&&\multicolumn{4}{c|}{\textbf{TGIF}}&\multicolumn{4}{c|}{\textbf{Charades}}&\multicolumn{4}{c}{\textbf{YouCook2}}\\\cline{4-15}

\multicolumn{1}{l}{Method}&\multicolumn{1}{c}{IS}&\multicolumn{1}{c|}{Venue}
&\multicolumn{1}{c}{$R@1\uparrow$}
&\multicolumn{1}{c}{$R@5\uparrow$}
&\multicolumn{1}{c}{$MeanR\downarrow$}
&\multicolumn{1}{c}{$RSum\uparrow$}
&\multicolumn{1}{|c}{$R@1\uparrow$} 
&\multicolumn{1}{c}{$R@5\uparrow$}
&\multicolumn{1}{c}{$MeanR\downarrow$}
&\multicolumn{1}{c}{$RSum\uparrow$}
&\multicolumn{1}{|c}{$R@1\uparrow$} 
&\multicolumn{1}{c}{$R@5\uparrow$}
&\multicolumn{1}{c}{$MeanR\downarrow$}
&\multicolumn{1}{c}{$RSum\uparrow$}\\\hline
RIVRL~\cite{DBLP:journals/tcsv/DongWCQLHW22}&-& TCSVT'22& 12.2&26.8&-&74.4& -&-&-&-&-&-&-&-\\
X-Pool~\cite{gorti2022xpool} & - & CVPR'22 &-&-&-&-& 16.1&35.2&67.2&96.2&-&-&-&-\\
AME-Net~\cite{DBLP:journals/tomccap/HanCZWC22}&-&TOMM'22& -&-&-&-&-&-&-&- & 7.6&21.5&-&61.9 \\
EMCL-Net$^*$~\cite{DBLP:conf/nips/JinHLWGSC022} & - & NeurIPS'22& 13.2&27.3&327.5&75.3& 16.4&36.3&69.5& 100.0&11.8&28.8&121.6&80.4\\ 
RoME~\cite{DBLP:journals/corr/abs-2206-12845}& - & CoRR'22& -&-&-&-&-&-&-&-& 6.3&16.9&-&48.4\\
SwAMP~\cite{DBLP:conf/aistats/Kim23} &- &AISTATS'23& -&-&-&-&-&-& -&- & 9.4&24.9&-&69.6 \\
MuMUR~\cite{DBLP:journals/ir/MadasuASRTBL23}&-&Inf. Retr. J.'23& -&-&-&-& 16.6&37.5&52.7&104.1&-&-&-&-\\
UCoFiA$^*$~\cite{DBLP:conf/iccv/WangSCBB23} & - & ICCV'23 & 13.0&27.1&323.8&74.8&16.4&37.2&74.6&101.3&\textbf{12.4}&30.4&123.5&83.8\\
LTME~\cite{DBLP:conf/icassp/ChengKWQ024}&-&ICASSP'24& 10.6& 24.1&-&66.9& -&-&-&-&-&-& -&-\\
BiC-Net~\cite{DBLP:journals/tomccap/HanZSXCC24}&-&TOMM'24& -&-&-&-&-&-& -&-& 8.7&23.9&-&66.1  \\
\textbf{MAC-VR (ours)}&-& - & \textbf{14.2}&\textbf{28.8}&\textbf{294.4}&\textbf{79.7}&\textbf{17.8}&\textbf{40.0}&\textbf{55.0}&\textbf{109.2}&12.3&\textbf{30.7}&\textbf{106.7}&\textbf{85.3} \\\hline

EMCL-Net$^*$~\cite{DBLP:conf/nips/JinHLWGSC022} & DSL & NeurIPS'22 & 14.8&29.4&317.3&80.8& 18.2&38.2&63.1&105.7&12.6&31.7&104.5&86.8\\
\textbf{MAC-VR (ours)}&DSL& - &\textbf{16.2}&\textbf{31.2}&\textbf{284.3}&\textbf{86.4}&\textbf{19.5}&\textbf{42.1}&\textbf{52.4}&\textbf{116.0}&\textbf{13.1}&\textbf{32.1}&\textbf{99.3}&\textbf{88.6}\\\hline\hline
\end{tabular}}
\vspace*{-10pt}
\caption{Comparison with SOTA across datasets. $-$: unreported results. $^*$our reproduced results. IS: Inference Strategy. The highest performance in each block is marked in \textbf{bold}.}
\label{tab:comparison_sota}
\vspace*{-20pt}
\end{table*}
Another factor could be that we use the same foundation model settings (only 8 frames) to generate tags for every dataset. In the case of DiDeMo, we extract visual tags from the whole video and textual tags by concatenating individual captions. As a result, many frames may not match any labelled segments, adding irrelevant and detrimental information. We leave this investigation for future work.
\vspace{-16pt}
\subsection{Ablation Studies}
\vspace{-5pt}
\label{sec:ablation}

We provide ablations on MAC-VR using MSR-VTT-9k over: the number of tags and architecture design; choice of foundation models; and use of auxiliary captions vs. tags.
We provide Full tables and additional ablations in the supp. (Secs.~\ref{appendix:tables} and~\ref{appendix:ablations} respectively).\\
\begin{wrapfigure}{r}{0.6\textwidth}
    \vspace{-16pt}
    \centering
    \begin{minipage}[b]{0.28\textwidth}
        \centering
        \includegraphics[width=\linewidth]{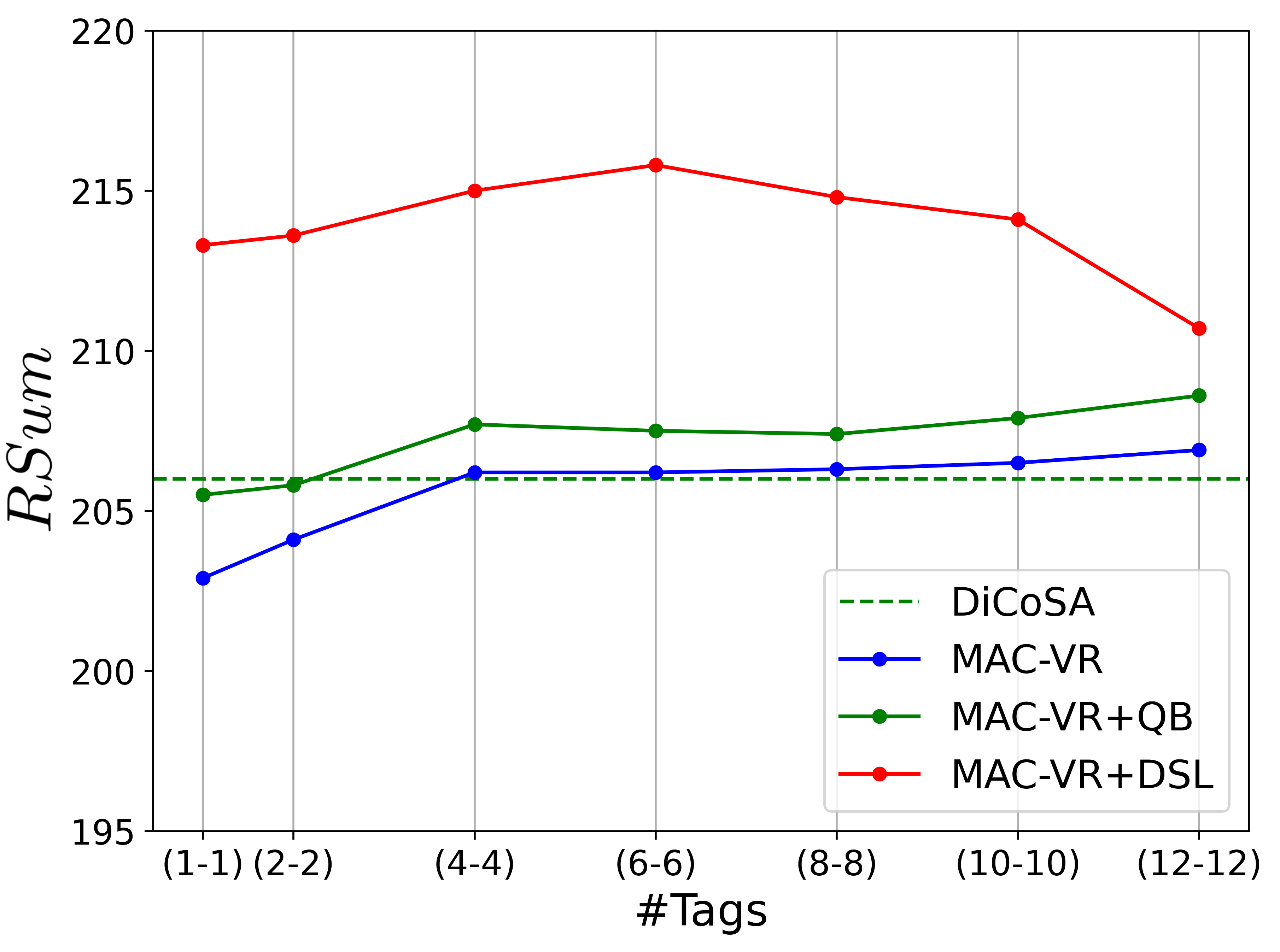}
    \end{minipage}
    \hfill
    \begin{minipage}[b]{0.28\textwidth}
        \centering
        \includegraphics[width=\linewidth]{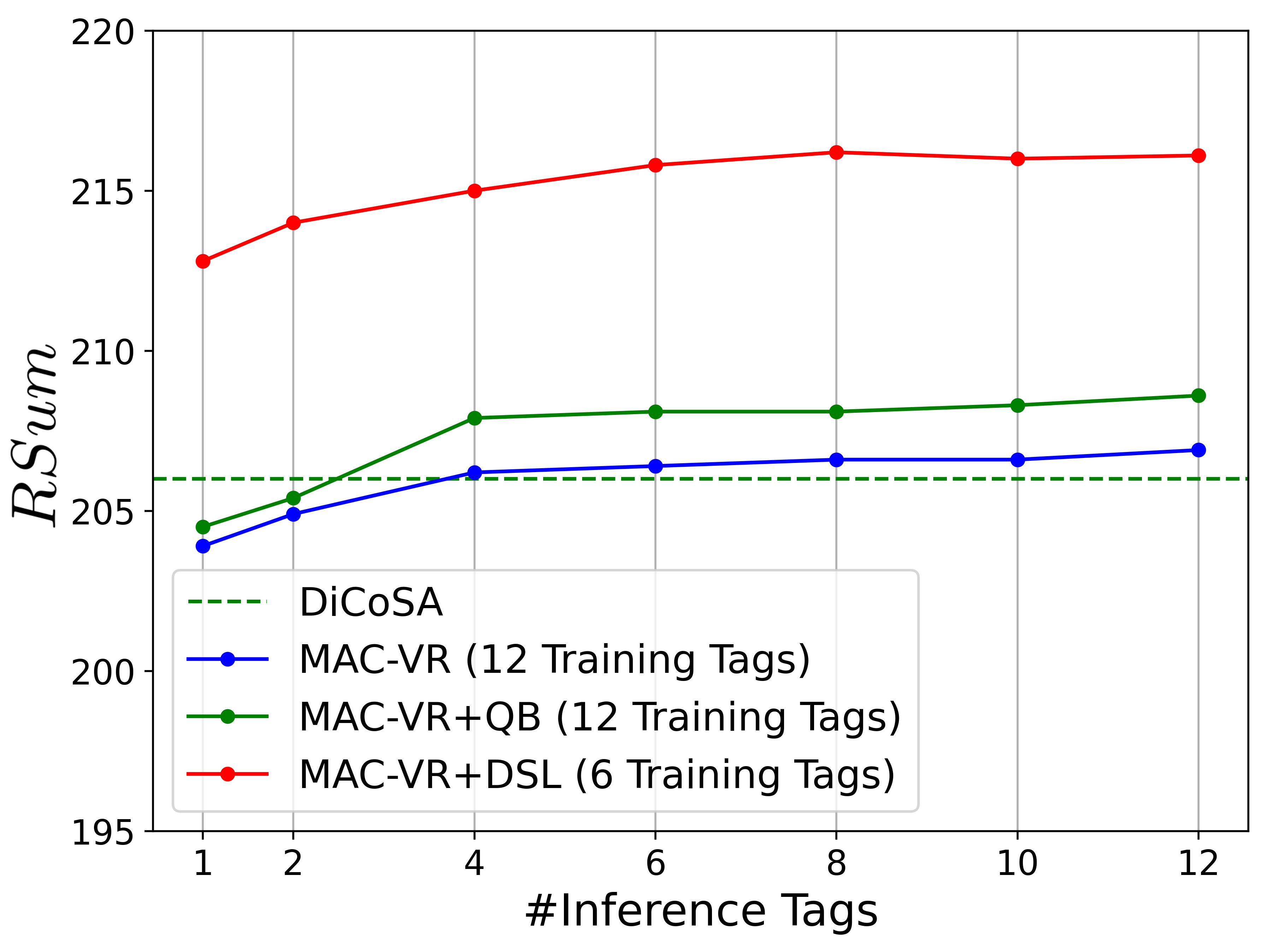}
    \end{minipage}
    \vspace{-10pt}
    \caption{Varying number of tags in Training/Inference (\textbf{left}) and Inference (\textbf{right}) across IS. $(n\text{-}m)$ indicates the num. of training tags $n$ and inference tags $m$.}
    \label{fig:different_number_tags}
\vspace{-10pt}
\end{wrapfigure}
\noindent \textbf{Num. of Tags in Training and Inference.} 
In Fig.~\ref{fig:different_number_tags} (left), we adjust only the number of tags in training and use the same value in inference. 
MAC-VR without IS and with QB increases in performance until reaching the best one with $12$ tags.
However, for DSL, $RSum$ increases until the best performance of $RSum=215.8$ with $6$ tags, then the value drops.
When varying the number of tags only in inference, as shown in Fig.~\ref{fig:different_number_tags} (right), $RSum$ increases rapidly and overcomes our baseline DiCoSA when using more than $2$ tags and gets the best performance with $RSum=216.2$ with $8$ tags using DSL. 
\begin{table}[ht!]
\centering

\begin{minipage}[ht!]{0.48\textwidth}
\centering
\resizebox{\linewidth}{!}{%
\begin{tabular}{l|cccc}
\multicolumn{1}{l|}{Method} & $R@1\uparrow$ & $R@5\uparrow$ & $MeanR\downarrow$ & $RSum\uparrow$ \\\hline
baseline & 52.1 & 77.3 & 12.9 & 215.3 \\
+VT      & 52.2 & 77.3 & 10.4 & 215.2 \\
+TT      & 52.0 & 77.6 & 10.4 & 215.6 \\
+VT+TT   & \textbf{53.2} & \textbf{77.7} & \textbf{10.0} & \textbf{216.2} \\\hline
\end{tabular}}
\caption{Ablation on architecture design.}
\label{tab:architecture}
\end{minipage}
\hfill
\begin{minipage}[ht!]{0.48\textwidth}
\centering
\resizebox{\linewidth}{!}{%
\begin{tabular}{cc|cccc}
\multicolumn{2}{c|}{Foundation Models}\\\cline{1-2}
\multicolumn{1}{c}{VT} & \multicolumn{1}{c|}{TT} &
\multicolumn{1}{c}{$R@1\uparrow$} & \multicolumn{1}{c}{$R@5\uparrow$} &
\multicolumn{1}{c}{$MeanR\downarrow$} & \multicolumn{1}{c}{$RSum\uparrow$} \\\hline
VL  & L2   & 52.0 & 77.5 & 10.4 & 214.4 \\
VL2 & L3.1 & \textbf{53.2} & \textbf{77.7} & \textbf{10.0} & \textbf{216.2} \\\hline\hline
\end{tabular}}
\caption{Ablation on foundation models. VL: Video-LLaMA. VL2: VideoLLaMA2. L2: Llama2. L3.1: Llama3.1.}
\label{tab:main_foundation_models}
\end{minipage}

\vspace{6pt}

\begin{minipage}[t!]{0.48\textwidth}
\centering
\resizebox{\linewidth}{!}{%
\begin{tabular}{ccc|cccc}
Auxiliary Input & Visual & Textual & $R@1\uparrow$ & $R@5\uparrow$ & $MeanR\downarrow$ & $RSum\uparrow$ \\\hline
Captions & Blip2 & PG    & 51.2 & 76.2 & 11.2 & 212.4 \\
Captions & Blip2 & L3.1  & 51.6 & 76.7 & 10.9 & 213.8 \\
Captions & VL2   & L3.1  & 50.4 & 75.9 & 11.5 & 211.0 \\
Tags     & VL2   & L3.1  & \textbf{53.2} & \textbf{77.7} & \textbf{10.0} & \textbf{216.2} \\\hline\hline
\end{tabular}}
\caption{Ablation on auxiliary inputs. PG: PEGASUS. L3.1: Llama3.1. VL2: VideoLLaMA2.}
\label{tab:main_aux_inputs}
\end{minipage}
\hfill
\begin{minipage}[ht!]{0.48\textwidth}
\centering
\resizebox{\linewidth}{!}{%
\begin{tabular}{cl|cccc}
& \multicolumn{1}{l|}{Inference Tags} & $R@1\uparrow$ & $R@5\uparrow$ & $MeanR\downarrow$ & $RSum\uparrow$ \\\hline
(i)   & original           & 53.2 & 77.7 & 10.0 & 216.2 \\
(ii)  & original wrong     & 50.4 & 76.5 & 10.8 & 211.4 \\
(iii) & self-correct       & \textbf{53.9} & \textbf{78.1} & \textbf{9.4} & \textbf{218.1} \\
(iv)  & self-correct wrong & 49.9 & 76.0 & 11.1 & 210.7 \\\hline
\end{tabular}}
\caption{Ablation on quality of auxiliary tags.}
\label{tab:quality_tag}
\end{minipage}

\vspace*{-14pt}
\end{table}

\noindent Best performance at inference is reported at $8$ tags with DSL and $12$ tags otherwise, showcasing that using more tags increases performance. The figure also shows that DSL consistently outperforms QB or no IS. Accordingly, we use DSL in other ablation studies.\\
\noindent\textbf{Choice of Tag Modality.} In Tab.~\ref{tab:architecture}, we ablate which modalities to extract tags from.
Using visual and textual tags individually gets results that are similar or slightly superior to baseline. When combined, all our metrics improve over the baseline.\\
\noindent\textbf{Choice of Foundation Models.} In Tab.~\ref{tab:main_foundation_models}, we ablate the use of different foundation models to extract visual and textual tags from a video and its caption.
We compare Video-LLaMA~\cite{damonlpsg2023videollama} against VideoLLaMA2~\cite{cheng2024videollama} and Llama2~\cite{touvron2023llama2} against Llama3.1~\cite{DBLP:journals/corr/abs-2407-21783} to extract visual and textual tags respectively.
Results show that all metrics improved using VideoLLaMA2 and Llama3.1 by $\Delta RSum=2.2$. Video-LLaMA and Llama2 tend to hallucinate tags more often, qualitative examples can be seen in Sec.~\ref{appendix:differences_tag_extraction} in supp.\\
\noindent\textbf{Auxiliary Captions vs Tags.} Tab.~\ref{tab:main_aux_inputs} shows that tags are more informative compared to additional captions extracted directly from the video or paraphrased from the caption. We use  Blip2~\cite{li2023blip} and VideoLLaMA2~\cite{cheng2024videollama} as video captioners and  PEGASUS~\cite{DBLP:conf/icml/ZhangZSL20} and Llama3.1~\cite{DBLP:journals/corr/abs-2407-21783} as paraphrasers, see supp. for details (Sec.~\ref{appendix:generate_new_captions}). 
Results show that tags outperforms additional captions -- including captions from the same models used to generate tags.\\
\noindent\textbf{Auxiliary Tags Quality.} Tab.~\ref{tab:quality_tag} evaluates tag quality during inference under four settings: (i) tags extracted from VLMs/LLMs; (ii) incorrect tags obtained by shuffling among samples; (iii) filtered tags, where a VLM/LLM filters the tag set to keep tags relevant to the video/caption (higher quality); and (iv) incorrect, filtered tags obtained by shuffling the filtered tags. Incorrect tags drop $RSum$ by $\text{-}4.8$ compared to (i), indicating that the auxiliary tags provide semantically relevant information. Using filtered tags improves $RSum$ by $\text{+}1.9$, suggesting that simple post-hoc filtering mitigates VLM/LLM hallucinations at the cost of additional filtering. Finally, wrong-filtered tags perform worst overall ($RSum=210.7$), reinforcing that curated tags are only beneficial when matched to the correct sample. We leave more sophisticated hallucination-mitigation approaches to future work.
\vspace{-16pt}
\section{Qualitative Results}
\begin{figure*}[t!]
\centering

\includegraphics[width=\linewidth]{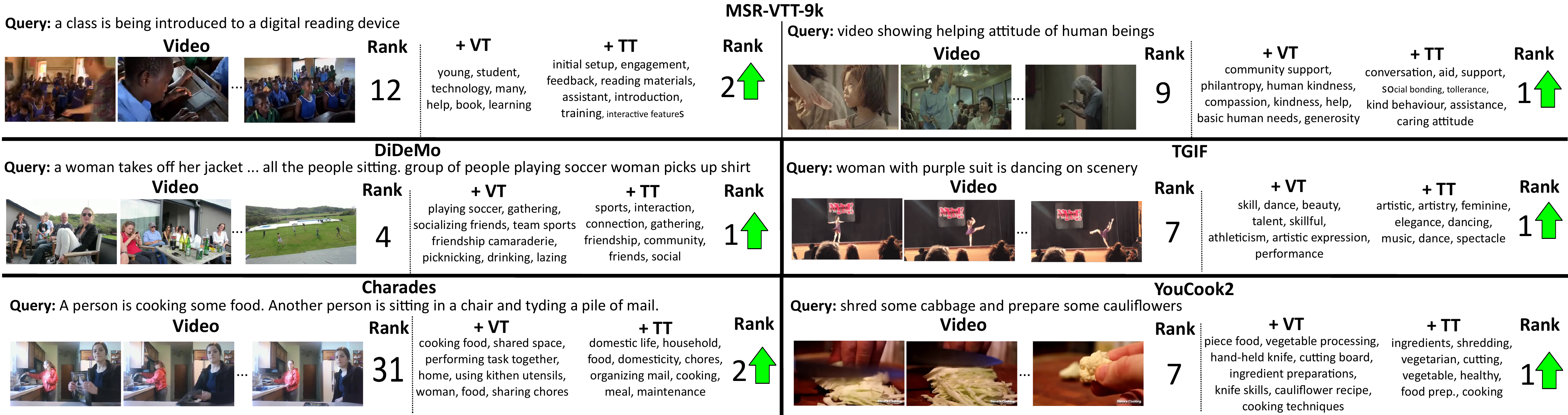}
    \vspace*{-20pt}
    \caption{Qual. results across all datasets comparing baseline/MAC-VR resp. left/right rank.}
    \vspace*{-6pt}
    \label{fig:qualitative_results}
\end{figure*}

\vspace{-8pt}
\label{sec:qualitative}
Fig.~\ref{fig:qualitative_results} shows qualitative results across all the datasets. We compare retrieved ranks of MAC-VR against our baseline without visual and textual tags. In general, both visual and textual tags add complementary information extracted from the video and the caption that directly help retrieval. For example, consider the query \textit{a class is being introduced to a digital reading device} and its video top right of Fig.~\ref{fig:qualitative_results}. 
Visual tags, \textit{student, technology, learning}, and textual tags, \textit{initial setup, introduction, training}, add additional information to describe the action shown in the video/described in the caption. We acknowledge there are cases where tags are harmful. Additional results, including failure cases are shown in Sec.~\ref{appendix:additional_qualitative} in supp.
\begin{figure*}[t!]
\centering
\begin{minipage}[b]{0.49\textwidth}
    \centering
    \includegraphics[width=\linewidth]{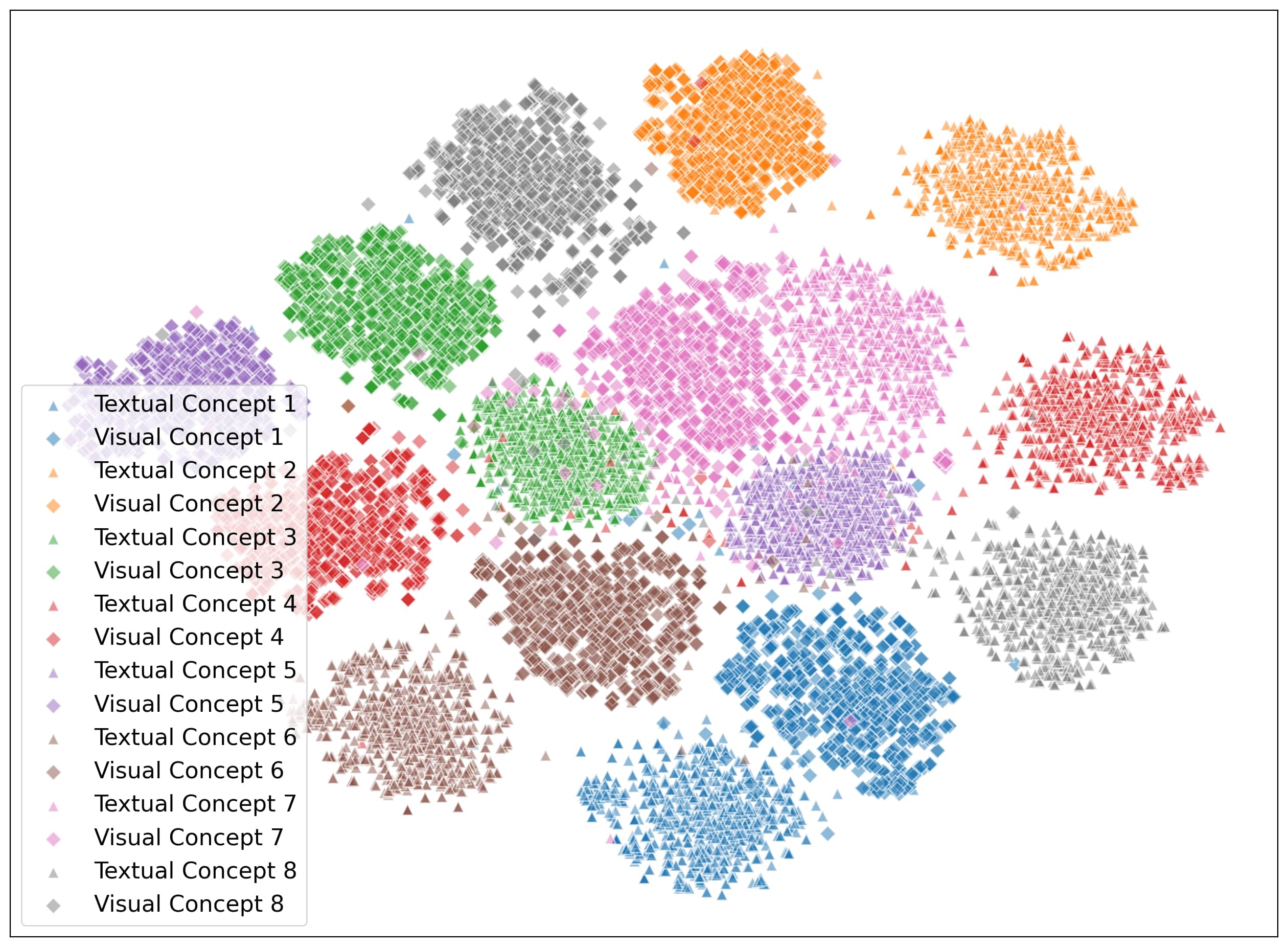}
\end{minipage}
\hfill
\begin{minipage}[b]{0.49\textwidth}
    \centering
    \includegraphics[width=\linewidth]{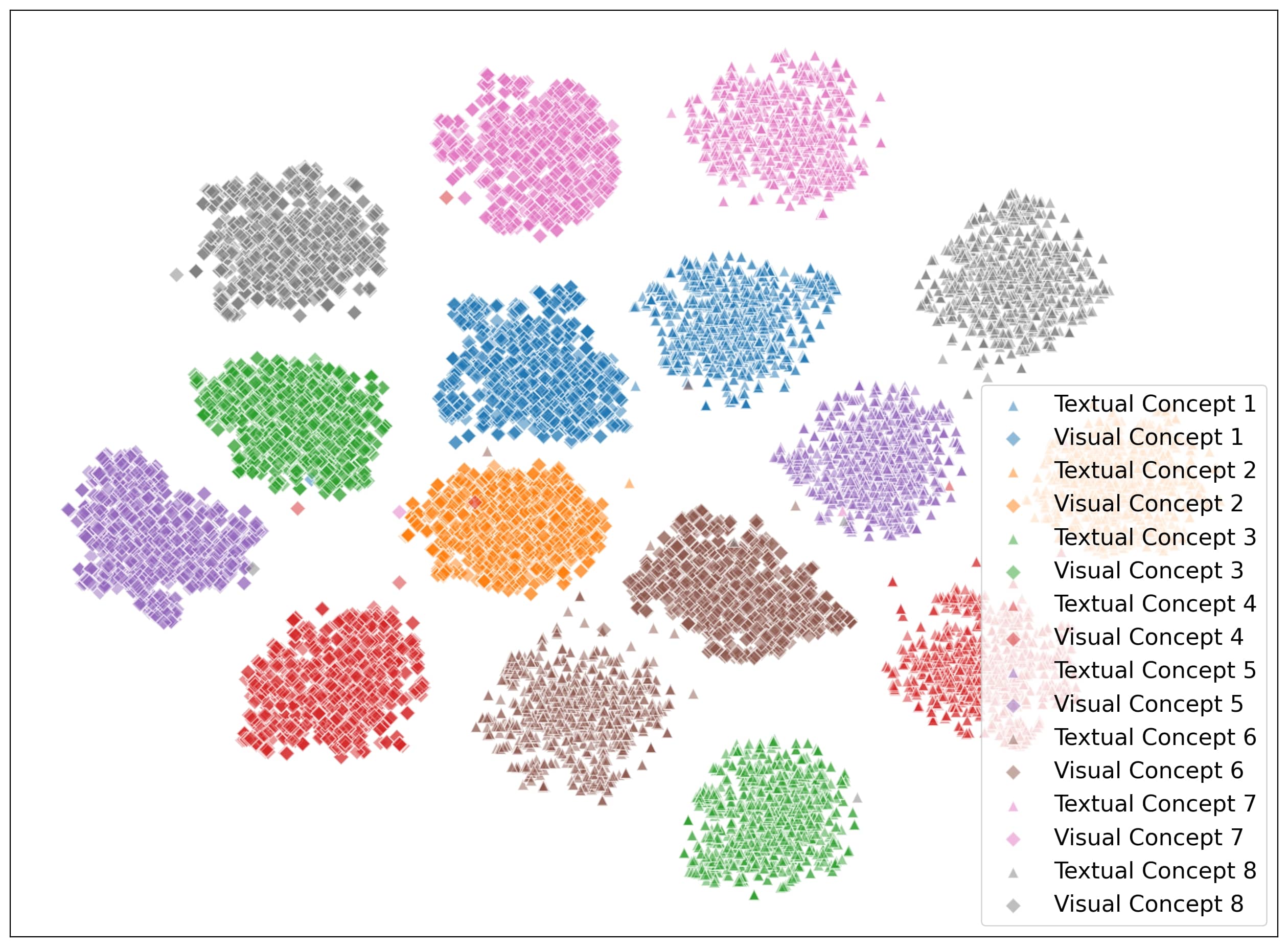}
\end{minipage}
\vspace{-8pt}
\caption{t-SNE plot of visual and textual concepts on MSR-VTT-9k without (\textbf{left}) and with using auxiliary modality-specific tags (\textbf{right}).}
\label{fig:tsne_plots}
\vspace*{-14pt}
\end{figure*}
Fig.~\ref{fig:tsne_plots} shows the t-SNE plot of MAC-VR of visual and textual concepts with/without auxiliary modality-specific tags on MSR-VTT-9k. We can see that the use of auxiliary modality-specific tags and our Alignment Loss $\mathcal{L_{A}}$ helps to better distinguish the different concepts and have better clusters in the t-SNE plot. Additional t-SNE plots are shown in the supp. 
\noindent\textbf{Discussion and Future Work.} We find that modality-specific tags are helpful for video retrieval, but they can be harmful for specific queries.
This could represent either correct but redundant tags or incorrectly extracted tags. Foundation models, i.e. VLMs and LLMs, often hallucinate the content of the output, generating factual reality or including fabricated information~\cite{DBLP:journals/corr/abs-2309-05922,DBLP:journals/corr/abs-2405-09589}.
Finally, we treat all the visual/textual tags with the same importance; it is possible that some words can be more or less discriminative than others based on uniqueness and/or relatedness to the video/caption. We leave this exploration for future work. 
\vspace{-16pt}
\section{Conclusion}
\vspace{-8pt}
In this work, we introduce the notion of visual and textual tags extracted by foundation models from a video and its caption respectively and use them to boost the video retrieval performance. We propose MAC-VR (Modality Auxiliary Concepts for Video Retrieval), where we incorporate modality-specific auxiliary tags, projected into disentangled auxiliary concepts. We use a new Alignment Loss to better align each modality with its auxiliary concepts.
We ablate our method to further show the benefit of using auxiliary modality-specific tags in video retrieval. Our results indicate, both qualitatively and by comparing to other approaches, that modality-specific tags help to decrease ambiguity in video retrieval on five video datasets.\\
\noindent \textbf{Acknowledgements} This work used public datasets and was supported by EPSRC Fellowship UMPIRE (EP/T004991/1) and EPSRC Program Grant Visual AI (EP/T028572/1).
\newpage

\bibliography{egbib}

\clearpage
\appendix

\section{Supplementary Contents}
 In the supplementary, we present further information about MAC-VR and further ablations over our design choices. In Sec.~\ref{appendix:inference_strategies} we introduce the Querybank Normalisation (QB)~\cite{DBLP:conf/cvpr/BogolinCJLA22} and Dual Softmax (DSL)~\cite{DBLP:journals/corr/abs-2109-04290}. Then, Sec.~\ref{appendix:tag_extract} provides additional details on the process of extracting tags from the foundation models, as well as further information about the extracted tags. In Sec.~\ref{appendix:datasets}, we describe in detail all the datasets used in the main paper. Then, in Sec.~\ref{appendix:tables}, we present the full tables shown in the main paper. Then, in Sec.~\ref{appendix:ablations}, we ablate additional architecture details of MAC-VR. After, we showcase the effect of using tags extracted from different foundation models in Sec.~\ref{appendix:differences_tag_extraction}. Then, we present a comparison to auxiliary captions in Sec.~\ref{appendix:generate_new_captions}, further t-SNE plots of MAC-VR in Sec.~\ref{appendix:additional_tsne}, and additional qualitative results in Sec.~\ref{appendix:additional_qualitative}.

\section{Inference Strategies}
\label{appendix:inference_strategies}

The QB strategy~\cite{DBLP:conf/cvpr/BogolinCJLA22} was introduced to mitigate the hubness problem of high-dimensional embedding spaces~\cite{DBLP:journals/jmlr/RadovanovicNI10}, where a small subset of samples tends to appear far more frequently among the k-nearest neighbours queries. 
To mitigate this, the similarities between embeddings are altered to minimise the influence of hubs. A set of samples (the querybank) is sampled from the queries within the training set and used as a probe to measure the hubness of the gallery at test time. More formally, given the vector of unnormalised similarities at test time, $S(v_j,t_i)$ and a probe matrix $P$, in which each row is a probe vector of similarities between the querybank and each element in the gallery, we re-weight the query-gallery similarities: $\eta_q= \text{QB}(S(v_j,t_i),P)$, where the $\text{QB}$ is the querybank normalisation function using Dynamic Inverted Softmax (DIS), introduced in~\cite{DBLP:conf/cvpr/BogolinCJLA22}.

The second commonly used inference strategy, the DSL strategy~\cite{DBLP:journals/corr/abs-2109-04290}, introduces an intrinsic prior of each pair in a batch to correct the similarity matrix and achieves the dual optimal match.
In practice, we modify the original $S(v_j,t_i)$ by multiplying it with a prior $r_{i,j}$. Therefore, we can define the new similarity matrix as 
$\hat{S}(v_j,t_i)=r_{i,j} S(v_j,t_i)$, where the prior is defined as $r_{i,j}=\frac{exp(\tau_r S(v_i,t_i))}{\sum_{j}exp(\tau_r S(v_i,t_j))}$, where $\tau_r$ is a temperature hyper-parameter to smooth the gradients. 
While this strategy can be used both in training and inference, it is now regularly used only during inference.
 
\section{Tag Extraction}
\label{appendix:tag_extract}

In Sec.~\ref{subsec:method:tags}, we have described how we extracted tags from a video and its corresponding caption, here we provide additional information and the prompts used to extract tags for the video and text modalities. 
For all the datasets except DiDeMo, we use the video and its corresponding caption to generate visual and textual tags, respectively. For DiDeMo, due to it being used for video-paragraph retrieval, we first concatenate the captions from the video to form the associated paragraph. Then, we generate tags from the entire video and paragraph.

\noindent The prompt used as input of VideoLLaMA2~\cite{cheng2024videollama} to extract visual tags is:
\begin{mdframed}[backgroundcolor=blue!10, linecolor=red, linewidth=2pt, roundcorner=10pt, shadow=true]

A general tag of an action is a fundamental and overarching idea that encapsulates the essential principles, commonalities, or recurrent patterns within a specific behaviour or activity, providing a higher-level understanding of the underlying themes and purpose associated with that action.\\ What are the top 10 general tags that capture the fundamental idea of this action? Give me a bullet list as output where each point is a general tag, and use one or two significant words per tag and do not give any explanation.

\end{mdframed}
\vspace{5pt}
The prompt for Llama3.1~\cite{DBLP:journals/corr/abs-2407-21783} to extract textual tags is: 
\begin{mdframed}[backgroundcolor=blue!10, linecolor=red, linewidth=2pt, roundcorner=10pt, shadow=true]
A chat between a curious user and an artificial intelligence assistant. The assistant gives helpful, detailed, and polite answers to the user's questions.\\
USER:
You are a conversational AI agent. You typically extract the general tags of an action.

A general tag of an action is a fundamental and overarching idea that encapsulates the essential principles, commonalities, or recurrent patterns within a specific behaviour or activity, providing a higher-level understanding of the underlying themes and purpose associated with that action.

Given the following action: 
1) \{\}

What are the top 10 general tags of the above action? Use one or two significant words per tag and do not give any explanation.

ASSISTANT: 

\end{mdframed}
\vspace{5pt}

Note that \{\} will be replaced with the caption.
We did not provide any examples in the prompts (i.e. in context learning) as we found that this led to the foundation models hallucinating these examples in the output. Rather, we found that the foundation models were able to generate reasonable outputs for both video and text.
\begin{figure*}[t!]
\centering
\includegraphics[width=0.82\linewidth]{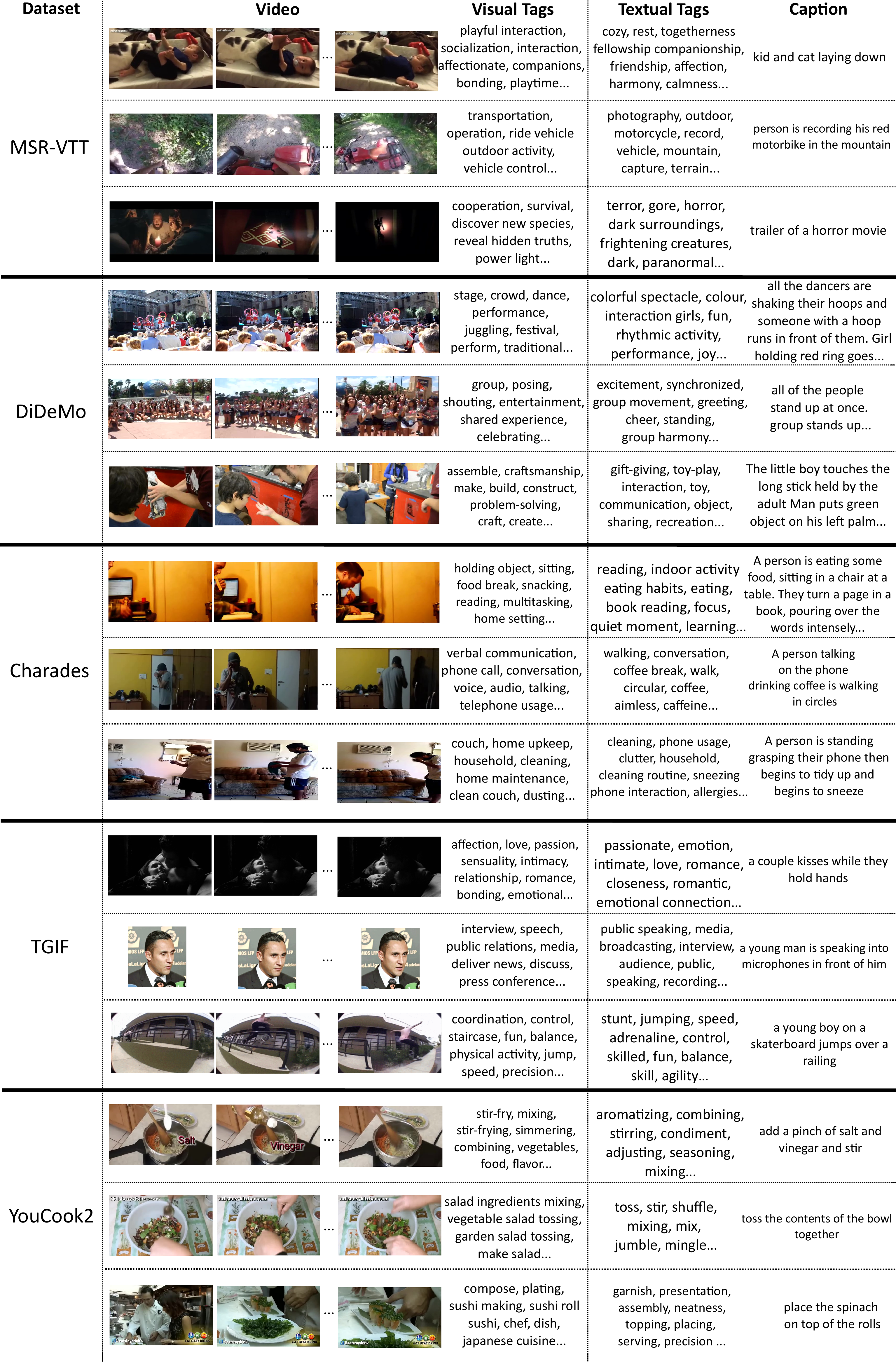}
    \caption{Additional examples of visual and textual tags across our datasets.}
    \label{fig:addtional_tags}
\end{figure*}

We do not use any strategy to avoid the foundation models hallucinating as spot-checking the results found them to be clean enough for our purposes. 
The only post-processing strategy we adopted was to clean the output of the models in order to get the corresponding tags: we remove punctuation; stopwords; extracted tags that contain a noun and a verb to avoid the presence of complete sentences as tags; and tags larger than $3$ words. Figure~\ref{fig:addtional_tags} illustrates additional tag examples for each dataset, while Table~\ref{tab:tags_info} summarises the number of unique visual and textual tags across all datasets.

\begin{table}[t!]
    \centering
    \resizebox{1\linewidth}{!}{%
\begin{tabular}{lcc|ccc|ccc}
&&&\multicolumn{3}{c|}{\#Unique Visual Tags (VT)}& \multicolumn{3}{c}{\#Unique Textual Tags (TT)}\\\cline{4-9}

\multicolumn{1}{l}{Datasets}&\multicolumn{1}{c}{\#Videos}&\multicolumn{1}{c|}{\# Captions}&\multicolumn{1}{c}{Train} &\multicolumn{1}{c}{Val}
&\multicolumn{1}{c|}{Test}
&\multicolumn{1}{c}{Train}
&\multicolumn{1}{c}{Val}
&\multicolumn{1}{c}{Test}\\\hline

MSR-VTT-9k~\cite{gabeur2020multi} &10,000&190,000&63,383&-&12,118&320,351&-&8,326\\
MSR-VTT-7k~\cite{DBLP:conf/iccv/MiechZATLS19} &8,010&141,200&52,309&-&12,118 &270,367&-&8,326\\
DiDeMo~\cite{DBLP:journals/corr/abs-1708-01641}&10,642&10,642&50,712&10,924&10,636&34,662 &9,234&8,266\\
TGIF~\cite{DBLP:conf/cvpr/LiSCTGJL16}&100,551&100,551&105,895&28,500&29,538&176,353&44,053&45,417\\
Charades~\cite{DBLP:conf/eccv/SigurdssonVWFLG16}&9,848&9,848&41,538&-&15,360&25,406&-&9,874\\
YouCook2~\cite{DBLP:conf/aaai/ZhouXC18}&13,829&13,829&33,786&-&16,457&26,035&-&12,227\\\hline\hline
\end{tabular}}

\caption{Statistical analysis of tags after extraction.}
\label{tab:tags_info}
\end{table}

\section{Datasets}
\label{appendix:datasets}
\noindent
\textbf{MSR-VTT}~\cite{xu2016msr} is commonly utilised for video retrieval. It comprises of 10K YouTube videos, each with $20$ captions. We utilise two commonly used training splits, with the same test set of 1K videos: the \textit{9k-Train} split~\cite{gabeur2020multi} and the \textit{7k-Train} split~\cite{DBLP:conf/iccv/MiechZATLS19}. \textbf{DiDeMo}~\cite{DBLP:journals/corr/abs-1708-01641} collects 10K Flickr videos annotated with paragraph captions. This dataset is evaluated in a video-paragraph retrieval manner~\cite{luo2022clip4clip}. 
\textbf{TGIF}~\cite{DBLP:conf/cvpr/LiSCTGJL16} is a large dataset that contains more than 100K animated short GIFs collected from Tumblr, and natural language sentences annotated via crowdsourcing.
We use the split proposed in~\cite{DBLP:conf/mm/LiX0CD19}.
\textbf{Charades}~\cite{DBLP:conf/eccv/SigurdssonVWFLG16} contains $9,848$ videos of $157$ action classes in total, where each video is associated with a caption. This dataset consists of fine-grained videos of recorded human actions. We use the standard training and test splits~\cite{DBLP:conf/eccv/LinLBB22}.
\textbf{YouCook2}~\cite{DBLP:conf/aaai/ZhouXC18} is a fine-grained dataset that contains 14K YouTube cooking videos that cover $89$ recipes. A textual sentence describes each video clip. Following~\cite{DBLP:conf/iccv/MiechZATLS19}, we evaluate this dataset on the validation set.
\begin{table*}[ht!]
\vspace*{-10pt}
\centering
\resizebox{\linewidth}{!}{
\begin{tabular}{lcc|cccccc|cccccc|cccccc}

&&&\multicolumn{6}{c|}{\textbf{MSR-VTT-9k} ($BS=128$, $N_v=12$)}&\multicolumn{6}{c|}{\textbf{MSR-VTT-7k} ($BS=128$, $N_v=12$)}&\multicolumn{6}{c}{\textbf{DiDeMo} ($BS=64$, $N_v=50$)}\\\cline{4-21}

\multicolumn{1}{l}{Method}&\multicolumn{1}{c}{IS}&\multicolumn{1}{c|}{Venue}&\multicolumn{1}{c}{$R@1\uparrow$} &\multicolumn{1}{c}{$R@5\uparrow$}&\multicolumn{1}{c}{$R@10\uparrow$} &\multicolumn{1}{c}{$MR\downarrow$} 
&\multicolumn{1}{c}{$MeanR\downarrow$}
&\multicolumn{1}{c}{$RSum\uparrow$}
&\multicolumn{1}{|c}{$R@1\uparrow$} &\multicolumn{1}{c}{$R@5\uparrow$}&\multicolumn{1}{c}{$R@10\uparrow$} &\multicolumn{1}{c}{$MR\downarrow$} 
&\multicolumn{1}{c}{$MeanR\downarrow$}
&\multicolumn{1}{c}{$RSum\uparrow$}
&\multicolumn{1}{|c}{$R@1\uparrow$} &\multicolumn{1}{c}{$R@5\uparrow$}&\multicolumn{1}{c}{$R@10\uparrow$} &\multicolumn{1}{c}{$MR\downarrow$} 
&\multicolumn{1}{c}{$MeanR\downarrow$}
&\multicolumn{1}{c}{$RSum\uparrow$}\\\hline

DiCoSA*~\cite{DBLP:conf/ijcai/JinLCHWYLC23}&-&IJCAI'23& 47.2 &73.5 &83.0&\textbf{2}&12.9&203.7&44.4&\textbf{72.6}&81.2&\textbf{2}&\textbf{14.7}&198.2&41.2&71.3&81.3&\textbf{2}&\textbf{15.9}&193.8\\
\textbf{MAC-VR (ours)} &-& - & \textbf{48.8} & \textbf{74.4} & \textbf{83.7} & \textbf{2} & \textbf{12.3} & \textbf{206.9}&\textbf{45.3}&71.9&\textbf{81.8}&\textbf{2}&15.0&\textbf{199.0}&\textbf{43.4}&\textbf{72.5}&\textbf{82.3}&\textbf{2}&16.9&\textbf{198.4}\\\hline

DiCoSA*~\cite{DBLP:conf/ijcai/JinLCHWYLC23}&QB&IJCAI'23& 48.0 &74.6&\textbf{84.3}&\textbf{2}&12.9&206.9&44.7&\textbf{73.3}&81.3&\textbf{2}&\textbf{14.3}&199.3&43.7&73.2&81.7&\textbf{2}&16.8&198.6\\
\textbf{MAC-VR (ours)}&QB& - & \textbf{49.3}& \textbf{75.9} & 83.5 & \textbf{2} & \textbf{12.3} & \textbf{208.7}&\textbf{45.6}&72.7&\textbf{82.1}&\textbf{2}&16.0&\textbf{200.4}&\textbf{45.5}&\textbf{74.8}&\textbf{82.3}&\textbf{2}&\textbf{16.2}&\textbf{202.6}\\\hline

DiCoSA*~\cite{DBLP:conf/ijcai/JinLCHWYLC23}&DSL&IJCAI'23& 52.1& 77.3&\textbf{85.9}&\textbf{1}&12.9&215.3&49.2&73.3&82.8&\textbf{2}&\textbf{12.6}&205.3&47.3&75.7&83.8&2&\textbf{14.2}&206.8\\
\textbf{MAC-VR (ours)}&DSL& - &\textbf{53.2}  & \textbf{77.7} & 85.3 & \textbf{1} & \textbf{10.0}& \textbf{216.2}&\textbf{49.8}&\textbf{74.6}&\textbf{84.3}&\textbf{2}&12.7&\textbf{208.7}&\textbf{50.2}&\textbf{76.2}&\textbf{84.2}&\textbf{1}&15.1&\textbf{209.6}\\\hline\hline
\end{tabular}}

\vspace*{2pt}

\resizebox{\linewidth}{!}{
\begin{tabular}{lcc|cccccc|cccccc|cccccc}
&&&\multicolumn{6}{c|}{\textbf{TGIF} ($BS=128$, $N_v=12$)}&\multicolumn{6}{c|}{\textbf{Charades} ($BS=128$, $N_v=12$)}&\multicolumn{6}{c}{\textbf{YouCook2} ($BS=128$, $N_v=12$)}\\\cline{4-21}
\multicolumn{1}{l}{Method}&\multicolumn{1}{c}{IS}&\multicolumn{1}{c|}{Venue}&\multicolumn{1}{c}{$R@1\uparrow$} &\multicolumn{1}{c}{$R@5\uparrow$}&\multicolumn{1}{c}{$R@10\uparrow$} &\multicolumn{1}{c}{$MR\downarrow$} 
&\multicolumn{1}{c}{$MeanR\downarrow$}
&\multicolumn{1}{c}{$RSum\uparrow$}
&\multicolumn{1}{|c}{$R@1\uparrow$} &\multicolumn{1}{c}{$R@5\uparrow$}&\multicolumn{1}{c}{$R@10\uparrow$} &\multicolumn{1}{c}{$MR\downarrow$} 
&\multicolumn{1}{c}{$MeanR\downarrow$}
&\multicolumn{1}{c}{$RSum\uparrow$}
&\multicolumn{1}{|c}{$R@1\uparrow$} &\multicolumn{1}{c}{$R@5\uparrow$}&\multicolumn{1}{c}{$R@10\uparrow$} &\multicolumn{1}{c}{$MR\downarrow$} 
&\multicolumn{1}{c}{$MeanR\downarrow$}
&\multicolumn{1}{c}{$RSum\uparrow$}\\\hline

DiCoSA*~\cite{DBLP:conf/ijcai/JinLCHWYLC23}&-&IJCAI'23& 14.0 &28.5 &36.1&30&296.7&78.6&16.7&35.9&47.1&12&61.7&99.7&11.9&29.2&40.3&18&117.9&81.4\\
\textbf{MAC-VR (ours)} &-& - & \textbf{14.2} & \textbf{28.8} & \textbf{36.7} & \textbf{28} & \textbf{294.4} &\textbf{79.7}&\textbf{17.8}&\textbf{40.0}&\textbf{51.4}&\textbf{10}&\textbf{55.0}&\textbf{109.2}&\textbf{12.3}&\textbf{30.7}&\textbf{42.3}&\textbf{16}&\textbf{106.7}&\textbf{85.3}\\\hline

DiCoSA*~\cite{DBLP:conf/ijcai/JinLCHWYLC23}&QB&IJCAI'23&14.1&28.4&36.4&30&313.6&78.9&18.8&39.5&49.0&11&60.8&107.3&\textbf{12.9}&\textbf{31.5}&42.3&\textbf{16}&108.7&86.7\\
\textbf{MAC-VR (ours)}&QB& - & \textbf{14.8} & \textbf{29.7} & \textbf{37.5} & \textbf{28} & \textbf{289.9} & \textbf{82.0} & \textbf{19.4} & \textbf{40.5} & \textbf{52.2} & \textbf{10} & \textbf{55.4} & \textbf{112.1} & \textbf{12.9} & 31.4 & \textbf{43.4} & \textbf{16} & \textbf{107.1} & \textbf{87.7}\\\hline

DiCoSA*~\cite{DBLP:conf/ijcai/JinLCHWYLC23}&DSL&IJCAI'23&15.5&30.1&37.8&28&290.6&83.4&18.2&39.2&49.1&11&58.9&106.5&12.8&\textbf{32.2}&43.0&\textbf{15}&103.8&88.0\\
\textbf{MAC-VR (ours)}&DSL& - & \textbf{16.2} & \textbf{31.2} & \textbf{39.0} & \textbf{25} & \textbf{284.3} & \textbf{86.4} & \textbf{19.5} & \textbf{42.1} & \textbf{54.4} & \textbf{8} & \textbf{52.4} & \textbf{116.0} & \textbf{13.1} & 32.1 & \textbf{43.4} & \textbf{15} & \textbf{99.3} & \textbf{88.6}\\\hline\hline
\end{tabular}}
\vspace{-10pt}
\caption{Comparison with baseline trained by using same training parameters of MAC-VR. $^*$our reproduced results with same training parameters. IS: Inference Strategy. $BS$: Batch Size. $N_v$: Number of Frames. The highest performance in each block is marked in \textbf{bold}.}
\vspace*{-4pt}
\label{tab:supp_comparison_baseline}
\end{table*}

\begin{table*}[ht!]
\vspace*{-8pt}
\centering
\resizebox{\linewidth}{!}{
\begin{tabular}{lcc|cccccc|cccccc|cccccc}

&&&\multicolumn{6}{c|}{\textbf{MSR-VTT-9k}}&\multicolumn{6}{c|}{\textbf{MSR-VTT-7k}}&\multicolumn{6}{c}{\textbf{DiDeMo}}\\\cline{4-21}

\multicolumn{1}{l}{Method}&\multicolumn{1}{c}{IS}&\multicolumn{1}{c|}{Venue}&\multicolumn{1}{c}{$R@1\uparrow$} &\multicolumn{1}{c}{$R@5\uparrow$}&\multicolumn{1}{c}{$R@10\uparrow$} &\multicolumn{1}{c}{$MR\downarrow$} 
&\multicolumn{1}{c}{$MeanR\downarrow$}
&\multicolumn{1}{c}{$RSum\uparrow$}
&\multicolumn{1}{|c}{$R@1\uparrow$} &\multicolumn{1}{c}{$R@5\uparrow$}&\multicolumn{1}{c}{$R@10\uparrow$} &\multicolumn{1}{c}{$MR\downarrow$} 
&\multicolumn{1}{c}{$MeanR\downarrow$}
&\multicolumn{1}{c}{$RSum\uparrow$}
&\multicolumn{1}{|c}{$R@1\uparrow$} &\multicolumn{1}{c}{$R@5\uparrow$}&\multicolumn{1}{c}{$R@10\uparrow$} &\multicolumn{1}{c}{$MR\downarrow$} 
&\multicolumn{1}{c}{$MeanR\downarrow$}
&\multicolumn{1}{c}{$RSum\uparrow$}\\\hline
CLIP4Clip~\cite{DBLP:journals/ijon/LuoJZCLDL22}&-&Neurocomp.'22&44.5&71.4&81.6&\textbf{2}&15.3&197.5&42.1&71.9&81.4&\textbf{2}&16.2&195.4&43.4&70.2&80.6&2&17.5&194.2\\
CenterCLIP~\cite{DBLP:conf/sigir/ZhaoZWY22}&-&SIGIR'22  & 44.2 & 71.6 & 82.1 & \textbf{2} & 15.1 &197.9&43.7&71.3&80.8&\textbf{2}&16.9&195.8& -&-&-&-&-&-\\
X-Pool~\cite{gorti2022xpool} &-& CVPR'22 & 46.9 & 72.8 & 82.2 & \textbf{2} & 14.3&201.9 &43.9&72.5&82.3&\textbf{2}&\textbf{14.6}&198.7&-&- & - & - & - & -\\
TS2-Net~\cite{DBLP:conf/eccv/LiuXXCJ22} &-&ECCV'22  & 47.0 & 74.5 & 83.8 & \textbf{2} & 13.0&205.3&-&-&-&-&-&-& 41.8 & 71.6 & 82.0 & 2 & 14.8&195.4\\
EMCL-Net~\cite{DBLP:conf/nips/JinHLWGSC022} &-& NeurIPS'22 & 46.8 & 73.1 & 83.1 & \textbf{2} & - &203.0&-&-&-&-&-&-& 46.8 & 74.3 & 83.1 & 2 & 12.3& 204.2 \\
EMCL-Net$^*$~\cite{DBLP:conf/nips/JinHLWGSC022} &-& NeurIPS'22 &-&-&-&-&-&-&45.2&71.4&81.2&\textbf{2}&15.0&197.8&44.1&71.7&80.3&2&15.1&197.8\\
VoP~\cite{Huang_2023_CVPR}&-&CVPR'23 & 44.6 & 69.9 & 80.3 & \textbf{2} & 16.3&194.8&42.7&68.2&79.3&\textbf{2}&15.9&190.2& 46.4 & 71.9 & 81.5 & 2 & 13.6&199.8\\
MuMUR~\cite{DBLP:journals/ir/MadasuASRTBL23}&-&Inf. Retr. J.'23&46.4&72.6&82.2&\textbf{2}&13.9&201.2&44.8&\textbf{72.0}&\textbf{82.5}&\textbf{2}&-&\textbf{199.3}&44.4&74.3&83.1&2&-&201.8\\
PiDRo~\cite{guan2023pidro} &-&ICCV'23& 48.2 & 74.9 & 83.3 & \textbf{2} & 12.6&206.4&-&-&-&-&-&-& 48.6 & 75.9 & 84.4 & 2 & 11.8&208.9\\
HBI~\cite{jin2023video} &-&CVPR'23 & 48.6 & 74.6 & 83.4 & \textbf{2} & \textbf{12.0}&206.6&-&-&-&-&-&-& 46.9 & 74.9 & 82.7 & 2 & 12.1 &204.5\\
DiffusionRet~\cite{jin2023diffusionret} &-& ICCV'23 & 49.0 & 75.2 & 82.7 & \textbf{2} & 12.1&206.9&-&-&-&-&-&-& 46.7 & 74.7 & 82.7 & 2 & 14.3&204.1  \\
Prompt Switch~\cite{deng2023prompt} &-& ICCV'23& 47.8 & 73.9 & 82.2 & - & 14.4&203.9&-&-&-&-&-&-& - & - & - & - & -&-\\
Cap4Video~\cite{DBLP:conf/cvpr/WuLFWO23} &-&CVPR'23& 49.3 & 74.3 & 83.8 & \textbf{2} & \textbf{12.0}&\textbf{207.4}&-&-&-&-&-&-& \textbf{52.0} & \textbf{79.4} & \textbf{87.5} & \textbf{1} & \textbf{10.5} &\textbf{218.9}\\
UCoFiA~\cite{DBLP:conf/iccv/WangSCBB23}&-& ICCV'23& \textbf{49.4} & 72.1 & 83.5 & \textbf{2} & 12.9&205.0&-&-&-&-&-&-& 46.5 & 74.8 & 84.4 & 2 & 13.4&205.7\\
UCoFiA$^*$~\cite{DBLP:conf/iccv/WangSCBB23}&-& ICCV'23&-&-&-&-&-&-&45.2&69.6&79.4&\textbf{2}&15.7&194.2&42.1&69.2&79.1&2&16.3&190.4\\
PAU~\cite{DBLP:conf/nips/LiSGZS23} &-& NeurIPS'23& 48.5 & 72.7 & 82.5 & \textbf{2} & 14.0 &203.7&-&-&-&-&-&-& 48.6 & 76.0 & 84.5 & 2 & 12.9&209.1\\
TABLE~\cite{chen2023tagging}&-& AAAI'23 & 47.1& 74.3 & 82.9 & \textbf{2} & 13.4&204.3&-&-&-&-&-&-& 47.9 & 74.0 & 82.1 & 2 & 14.3&204.0\\
UATVR~\cite{DBLP:conf/iccv/FangWLZS0SJW23}&-& ICCV'23 & 47.5& 73.9 & 83.5 & \textbf{2} & 12.3&204.9&-&-&-&-&-&-& 43.1 & 71.8 & 82.3 & 2 & 15.1&197.2\\
TeachCLIP~\cite{tian2024holistic} &-& CVPR'24& 46.8 & 74.3 & - & - & - &-&-&-&-&-&-&-& 43.7 & 71.2 & - & - & -&- \\
MV-Adapter~\cite{jin2024mv} &-&CVPR'24& 46.2 & 73.2 & 82.7 & - & - &202.1&-&-&-&-&-&-& 44.3 & 72.1 & 80.5 & - & - &196.9\\
BiC-Net~\cite{DBLP:journals/tomccap/HanZSXCC24}&-&TOMM'24&39.4&\textbf{75.5}&\textbf{86.7}&\textbf{2}&-&201.6&32.8&68.2&82.4&3&-&183.4&-&-&-&-&-&-\\
RAP~\cite{DBLP:conf/acl/CaoTHJ0L0LY024}&-&ACL'24&44.8&71.4&81.5&-&14.4&197.7&-&-&-&-&-&-&42.6&70.4&79.6&-&18.0&192.6\\
\textbf{MAC-VR (ours)} &-& - & 48.8 & 74.4 & 83.7 & \textbf{2} & 12.3 &206.9&\textbf{45.3}&71.9&81.8&\textbf{2}&15.0&199.0&43.4&72.7&82.3&2&16.9&198.4\\\hline

QB-Norm~\cite{DBLP:conf/cvpr/BogolinCJLA22} &QB&CVPR'22& 47.2 & 73.0 & 83.0 & \textbf{2} & -&203.2&-&-&-&-&-&-& 43.3 & 71.4 & 80.8 & \textbf{2} & -&195.5\\
DiCoSA~\cite{DBLP:conf/ijcai/JinLCHWYLC23} &QB& IJCAI'23& 47.5 & 74.7 & \textbf{83.8} & \textbf{2} & 13.2&206.0&-&-&-&-&-&-& 45.7 & 74.6 & \textbf{83.5} & \textbf{2} & \textbf{11.7}&203.8 \\
DiffusionRet~\cite{jin2023diffusionret} &QB& ICCV'23 & 48.9 & 75.2 & 83.1 & \textbf{2} & \textbf{12.1}&207.2&-&-&-&-&-&-& \textbf{48.9} & \textbf{75.5} & 83.3 & \textbf{2} & 14.1  &\textbf{207.7}\\
\textbf{MAC-VR (ours)}&QB& - & \textbf{49.3}& \textbf{75.9} & 83.5 & \textbf{2} & 12.3 &\textbf{208.7}&\textbf{45.6}&\textbf{72.7}&\textbf{82.1}&\textbf{2}&\textbf{16.0}&\textbf{200.4}&45.5&74.8&82.3&\textbf{2}&16.2&202.6\\\hline

EMCL-Net~\cite{DBLP:conf/nips/JinHLWGSC022} &DSL& NeurIPS'22 & 51.6 & 78.1 & 85.3 & \textbf{1} & - &215.0&-&-&-&-&-&-& - & - & - & - & -& -\\
EMCL-Net$^*$~\cite{DBLP:conf/nips/JinHLWGSC022} &DSL& NeurIPS'22 &-&-&-&-&-&-&-&-&-&-&-&-&47.6&73.5&82.8&2&\textbf{11.9}&203.9\\
TS2-Net~\cite{DBLP:conf/eccv/LiuXXCJ22} &DSL&ECCV'22 & 51.1 & 76.9 & 85.6 & \textbf{1} & 11.7 &213.6&-&-&-&-&-&-& 47.4 & 74.1 & 82.4 & 2 & 12.9&203.9\\
TABLE~\cite{chen2023tagging}&DSL& AAAI'23& 52.3 & 78.4 & 85.2 & \textbf{1} & 11.4&215.9&-&-&-&-&-&-& 49.1 & \textbf{75.6} & 82.9 & 2 & 14.8 &207.6\\
UATVR~\cite{DBLP:conf/iccv/FangWLZS0SJW23}&DSL& ICCV'23 & 49.8& 76.1 & 85.5 & 2 & 12.3&211.4&-&-&-&-&-&-& - & - & - & - & - &-\\
RAP~\cite{DBLP:conf/acl/CaoTHJ0L0LY024}&DSL& ACL'24&-&-&-&-&-&-&-&-&-&-&-&-&47.1&74.1&82.4&-&13.9&203.6\\
\textbf{MAC-VR (ours)}&DSL& -&\textbf{53.2}  & 77.7 & 85.3 & \textbf{1} & \textbf{10.0}&\textbf{216.2}&\textbf{49.8}&\textbf{74.6}&\textbf{84.3}&\textbf{2}&\textbf{12.7}&\textbf{208.7}&\textbf{50.2}&75.2&\textbf{84.2}&\textbf{1}&15.1&\textbf{209.6}\\\hline\hline
\end{tabular}}

\vspace*{2pt}

\resizebox{\linewidth}{!}{
\begin{tabular}{lcc|cccccc|cccccc|cccccc}

&&&\multicolumn{6}{c|}{\textbf{TGIF}}&\multicolumn{6}{c|}{\textbf{Charades}}&\multicolumn{6}{c}{\textbf{YouCook2}}\\\cline{4-21}

\multicolumn{1}{l}{Method}&\multicolumn{1}{c}{IS}&\multicolumn{1}{c|}{Venue}&\multicolumn{1}{c}{$R@1\uparrow$} &\multicolumn{1}{c}{$R@5\uparrow$}&\multicolumn{1}{c}{$R@10\uparrow$} &\multicolumn{1}{c}{$MR\downarrow$} 
&\multicolumn{1}{c}{$MeanR\downarrow$}
&\multicolumn{1}{c}{$RSum\uparrow$}
&\multicolumn{1}{|c}{$R@1\uparrow$} &\multicolumn{1}{c}{$R@5\uparrow$}&\multicolumn{1}{c}{$R@10\uparrow$} &\multicolumn{1}{c}{$MR\downarrow$} 
&\multicolumn{1}{c}{$MeanR\downarrow$}
&\multicolumn{1}{c}{$RSum\uparrow$}
&\multicolumn{1}{|c}{$R@1\uparrow$} &\multicolumn{1}{c}{$R@5\uparrow$}&\multicolumn{1}{c}{$R@10\uparrow$} &\multicolumn{1}{c}{$MR\downarrow$} 
&\multicolumn{1}{c}{$MeanR\downarrow$}
&\multicolumn{1}{c}{$RSum\uparrow$}\\\hline
RIVRL~\cite{DBLP:journals/tcsv/DongWCQLHW22}&-&TCSVT'22& 12.2&26.8&35.4&30&-&74.4& -&-&-&-&-&-&-&-&-&-&-&- \\
X-Pool~\cite{gorti2022xpool} & - & CVPR'22 & -&-&-&-&-&-& 16.1&35.2&44.9&14&67.2&96.2&-&-&-&-&-&-\\
AME-Net~\cite{DBLP:journals/tomccap/HanCZWC22}&-&TOMM'22& -&-&-&-&-&-& -&-&-&-&-&- & 7.6&21.5&32.8&28&-&61.9 \\
EMCL-Net$^*$~\cite{DBLP:conf/nips/JinHLWGSC022} & - & NeurIPS'22 & 13.2&27.3&34.8&36&327.5&75.3& 16.4&36.3&47.3&13&69.5& 100.0&11.8&28.8&39.8&19&121.6&80.4\\ 
RoME~\cite{DBLP:journals/corr/abs-2206-12845}& - & CoRR'22 & -&-&-&-&-&-& -&-&-&-&-&-& 6.3&16.9&25.2&53&-&48.4\\
SwAMP~\cite{DBLP:conf/aistats/Kim23} &- &AISTATS'23& -&-&-&-&-&-& -&-&-&-&-&- & 9.4&24.9&35.3&22&-&69.6 \\
MuMUR~\cite{DBLP:journals/ir/MadasuASRTBL23}&-&Inf. Retr. J.'23& -&-&-&-&-&-& 16.6&37.5&50.0&\textbf{10}&52.7&104.1&-&-&-&-&-&- \\
UCoFiA$^*$~\cite{DBLP:conf/iccv/WangSCBB23} & - & ICCV'23& 13.0&27.1&34.7&35&323.8&74.8&16.4&37.2&47.7&12&74.6&101.3&\textbf{12.4}&30.4& 41.0&17&123.5&83.8\\
LTME~\cite{DBLP:conf/icassp/ChengKWQ024}&-&ICASSP'24& 10.6& 24.1& 32.2&-&-&66.9& -&-&-&-&-&-& -&-&-&-&-&-\\
BiC-Net~\cite{DBLP:journals/tomccap/HanZSXCC24}&-&TOMM'24& -&-&-&-&-&-& -&-&-&-&-&-& 8.7&23.9&33.5&31&-&66.1  \\
\textbf{MAC-VR (ours)}&-& - & \textbf{14.2}&\textbf{28.8}&\textbf{36.7}&\textbf{28}&\textbf{294.4}&\textbf{79.7}&\textbf{17.8}&\textbf{40.0}&\textbf{51.4}&\textbf{10}&\textbf{55.0}&\textbf{109.2}&12.3&\textbf{30.7}&\textbf{42.3}&\textbf{16}&\textbf{106.7}&\textbf{85.3} \\\hline

EMCL-Net$^*$~\cite{DBLP:conf/nips/JinHLWGSC022} & DSL & NeurIPS'22 & 14.8&29.4&36.6&31&317.3&80.8& 18.2&38.2&49.3&11&63.1&105.7&12.6&31.7&42.5&17&104.5&86.8\\
\textbf{MAC-VR (ours)}&DSL& -&\textbf{16.2}&\textbf{31.2}&\textbf{39.0}&\textbf{25}&\textbf{284.3}&\textbf{86.4}&\textbf{19.5}&\textbf{42.1}&\textbf{54.4}&\textbf{8}&\textbf{52.4}&\textbf{116.0}&\textbf{13.1}&\textbf{32.1}&\textbf{43.4}&\textbf{15}&\textbf{99.3}&\textbf{88.6}\\\hline\hline
\end{tabular}}
\vspace*{-10pt}
\caption{Comparison with SOTA across all the datasets. $-$: unreported results. $^*$our reproduced results. IS: Inference Strategy. The highest performance in each block is marked in \textbf{bold}.}
\label{tab:supp_comparison_sota}
\vspace*{-10pt}
\end{table*}

\section{Full Tables of Experiments}
\label{appendix:tables}
Tab.~\ref{tab:supp_comparison_baseline} details the results presented in Fig.~\ref{fig:comparison_baseline} and reports Recall at $L = 1, 5, 10$ ($R@L$), median rank $MR$, mean rank ($MeanR$) and $RSum$ for all the datasets when using all the inference strategies. Similarly, Tab.~\ref{tab:supp_comparison_sota} reports the full table of Tab.~\ref{tab:comparison_sota}. Tab.~\ref{tab:supp_architecture} is the full table of the ablation Tab.~\ref{tab:architecture} in the main paper. Tabs~\ref{tab:supp_architecture} to~\ref{tab:sup_quality_tag} shows the full tables with all the metrics of the ablation experiments in Sec.~\ref{sec:ablation} of the main paper.

\begin{table}[ht!]
    \centering
\begin{tabular}{l|cccccc}
\multicolumn{1}{l|}{Method}
&\multicolumn{1}{c}{$R@1\uparrow$} &\multicolumn{1}{c}{$R@5\uparrow$}&\multicolumn{1}{c}{$R@10\uparrow$} &\multicolumn{1}{c}{$MR\downarrow$} 
&\multicolumn{1}{c}{$MeanR\downarrow$}
&\multicolumn{1}{c}{$RSum\uparrow$} \\\hline
baseline & 52.1 & 77.3 &  85.9& \textbf{1} & 12.9& 215.3\\
+VT& 52.2 &   77.3  & 85.7& \textbf{1} & 10.4& 215.2 \\
+TT& 52.0 & 77.6 & \textbf{86.0} & \textbf{1} & 10.4 & 215.6\\ 
+VT+TT& \textbf{53.2} & \textbf{77.7} & 85.3 & \textbf{1} & \textbf{10.0}& \textbf{216.2}\\\hline\hline
\end{tabular}
\caption{Ablation on architecture design.}
\label{tab:supp_architecture}
\vspace{-10pt}
\end{table}

\begin{table}[ht!]
    \centering
\resizebox{0.8\linewidth}{!}{%
\begin{tabular}{ccc|cccccc}

\multicolumn{1}{c}{Auxiliary Input}&\multicolumn{1}{c}{Visual} &\multicolumn{1}{c|}{Textual} &\multicolumn{1}{c}{$R@1\uparrow$} &\multicolumn{1}{c}{$R@5\uparrow$}&\multicolumn{1}{c}{$R@10\uparrow$} &\multicolumn{1}{c}{$MR\downarrow$} 
&\multicolumn{1}{c}{$MeanR\downarrow$}
&\multicolumn{1}{c}{$RSum\uparrow$} \\\hline
Captions& Blip2&PG  & 51.2 & 76.2 & 85.0 & \textbf{1} & 11.2& 212.4\\
Captions& Blip2&L3.1  & 51.6 & 76.7 & \textbf{85.5} & \textbf{1} & 10.9 & 213.8\\
Captions& VL2&L3.1  & 50.4 & 75.9 & 84.7 & \textbf{1} & 11.5& 211.0\\
Tags& VL2&L3.1 
 &\textbf{53.2}  & \textbf{77.7} & 85.3 & \textbf{1} & \textbf{10.0} & \textbf{216.2}\\\hline\hline
\end{tabular}}
\caption{Ablation on auxiliary inputs. PG: PEGASUS. L3.1: Llama3.1. VL2: VideoLLaMA2.}
\label{tab:sup_captions}
    \vspace{-10pt}
\end{table}

\begin{table}[ht!]
    \centering
\resizebox{0.8\linewidth}{!}{%
\begin{tabular}{cc|cccccc}
\multicolumn{2}{c|}{Foundation Models}\\\cline{1-2}
\multicolumn{1}{c}{VT} &\multicolumn{1}{c|}{TT}&\multicolumn{1}{c}{$R@1\uparrow$} &\multicolumn{1}{c}{$R@5\uparrow$}&\multicolumn{1}{c}{$R@10\uparrow$} &\multicolumn{1}{c}{$MR\downarrow$} 
&\multicolumn{1}{c}{$MeanR\downarrow$}
&\multicolumn{1}{c}{$RSum\uparrow$} \\\hline
VL & L2 & 52.0 & 77.5  & 84.9 & \textbf{1} & 10.4& 214.4\\
VL2 & L3.1 & \textbf{53.2}  & \textbf{77.7} & \textbf{85.3} & \textbf{1} & \textbf{10.0}& \textbf{216.2}\\\hline\hline
\end{tabular}}
\caption{Ablation on foundation models. VL: Video-LLaMA. VL2: VideoLLaMA2. L2: Llama2. L3.1: Llama3.1.}
\label{tab:sup_foundation_models}
    \vspace{-10pt}
\end{table}

\begin{table}[ht!]
    \centering
\resizebox{0.8\linewidth}{!}{%
\begin{tabular}{cl|cccccc}
&\multicolumn{1}{l|}{Inference Tags}
& $R@1\uparrow$ & $R@5\uparrow$ & $R@10\uparrow$ & $MR\downarrow$& $MeanR\downarrow$ & $RSum\uparrow$ \\\hline
(i)&original & 53.2 & 77.7 & 85.3 & \textbf{1}& 10.0 & 216.2 \\
(ii)& original wrong & 50.4 & 76.5 & 84.5 & \textbf{1} &10.8 & 211.4 \\
(iii) & self-correct & \textbf{53.9} & \textbf{78.1} & \textbf{86.1} & \textbf{1}& \textbf{9.4} & \textbf{218.1} \\
(iv)& self-correct wrong & 49.9 & 76.0 & 84.8 & 2 & 11.1 & 210.7\\\hline
\end{tabular}}
\caption{Ablation on quality of auxiliary tags.}
\label{tab:sup_quality_tag}

\end{table}

\section{Additional Ablations}
\label{appendix:ablations}
In this section, we ablate some additional design choices of MAC-VR.
First, we ablate $K$ that controls the number of concepts/latent factors. We evaluate within the following range $K \in {4, 8, 16, 32}$. In Tab.~\ref{tab:ablation_k}, the performance improves reaching the best
result with $K = 8$ and then decreases. This shows that a small number of concepts limits the ability to leverage fine-grained information, whereas a larger value reduces the dimensionality ofceach concept limiting the discriminability of the concept itself.
\begin{table}[ht!]
    \centering
    \begin{tabular}{c|cccccc}
        $K$ &\multicolumn{1}{c}{$R@1\uparrow$} &\multicolumn{1}{c}{$R@5\uparrow$}&\multicolumn{1}{c}{$R@10\uparrow$} &\multicolumn{1}{c}{$MR\downarrow$} 
&\multicolumn{1}{c}{$MeanR\downarrow$}
&\multicolumn{1}{c}{$RSum\uparrow$}\\
        \hline
         4  & 52.4 & 77.1 & 85.1 & \textbf{1} & 10.2& 214.6 \\
         8  & \textbf{53.2} & \textbf{77.7} & \textbf{85.3} & \textbf{1} & \textbf{10.0} &\textbf{216.2}\\
         16 & 52.0 & 77.3 & 84.8 & \textbf{1} & 10.4& 214.1 \\
         32 & 50.7 & 77.1 & 84.5 & \textbf{1} & 10.3 & 212.3\\
        \hline\hline
    \end{tabular}
    \caption{Ablation on number of concepts $K$.}
    \label{tab:ablation_k}
    \vspace{3pt}
    \centering
    \begin{tabular}{cc|cccccc}
        $\alpha_C$& $\alpha_A$ &\multicolumn{1}{c}{$R@1\uparrow$} &\multicolumn{1}{c}{$R@5\uparrow$}&\multicolumn{1}{c}{$R@10\uparrow$} &\multicolumn{1}{c}{$MR\downarrow$} 
&\multicolumn{1}{c}{$MeanR\downarrow$}
&\multicolumn{1}{c}{$RSum\uparrow$}\\\hline
        0.5 & 1.0 &52.5 & 76.5 & 85.6 & \textbf{1} & 11.0 & 214.6\\
        1.0 & 1.0 &\textbf{53.2} & \textbf{77.7} & 85.3 & \textbf{1} & \textbf{10.0} & \textbf{216.2}\\
        2.0 & 1.0 &52.6 & 77.2 & \textbf{85.8} & \textbf{1} & 10.5 & 215.6\\
        5.0 & 1.0 &52.0 & 76.8 & 85.3 & \textbf{1} & 11.1 & 214.1\\\hline
        1.0 & 0.0 &52.1 & 77.3 & \textbf{85.9} & \textbf{1} & 12.9 & 215.3 \\
        1.0 & 0.5 &53.1 & 76.7 & 85.5 & \textbf{1} & \textbf{10.0} & 215.3\\
        1.0 & 1.0 &\textbf{53.2} & \textbf{77.7} & 85.3 & \textbf{1} & \textbf{10.0} & \textbf{216.2}\\
        1.0 & 2.0 &53.0 & 77.5 & 85.4 & \textbf{1} & 10.1 & 215.9\\
        1.0 & 5.0 &52.1 & 76.7 & 85.6 & \textbf{1} & 10.9 & 214.4\\
        1.0 & 10.0 &52.5 & 77.5 & 85.4 & \textbf{1} & 10.5 & 215.4\\\hline\hline
    \end{tabular}
        \caption{Ablation on loss parameters $\alpha_C$ and $\alpha_A$.}
    \label{tab:weight}
    \vspace{2pt}
    \vspace{-10pt}
\end{table}
In our training loss, we introduced two weight parameters $\alpha_C$ and $\alpha_A$ that control the importance of the contrastive loss $L_C$ to align the modality concepts and the proposed Alignment Loss $L_A$ to align the modality with the auxiliary modality concepts. 
We consider different values of $\alpha_C \in \{0.5, 1.0, 2.0, 5.0\}$ and $\alpha_A \in \{0.0, 0.5, 1.0, 2.0, 5.0, 10.0\}$. We find that the best $RSum$ is when $\alpha_C=1.0$ and $\alpha_A=1.0$ with $RSum=216.2$. The performance drops when $\alpha_C>2.0$ or $\alpha_A>2.0$, as shown in Tab.~\ref{tab:weight}.
These results show that both losses must contribute to the learning phase equally to better align visual and textual concepts and so improve the final retrieval performance.

\section{Differences between tags extracted from different foundations models}
\label{appendix:differences_tag_extraction}
\begin{figure*}[t!]
\centering
\includegraphics[width=\linewidth]{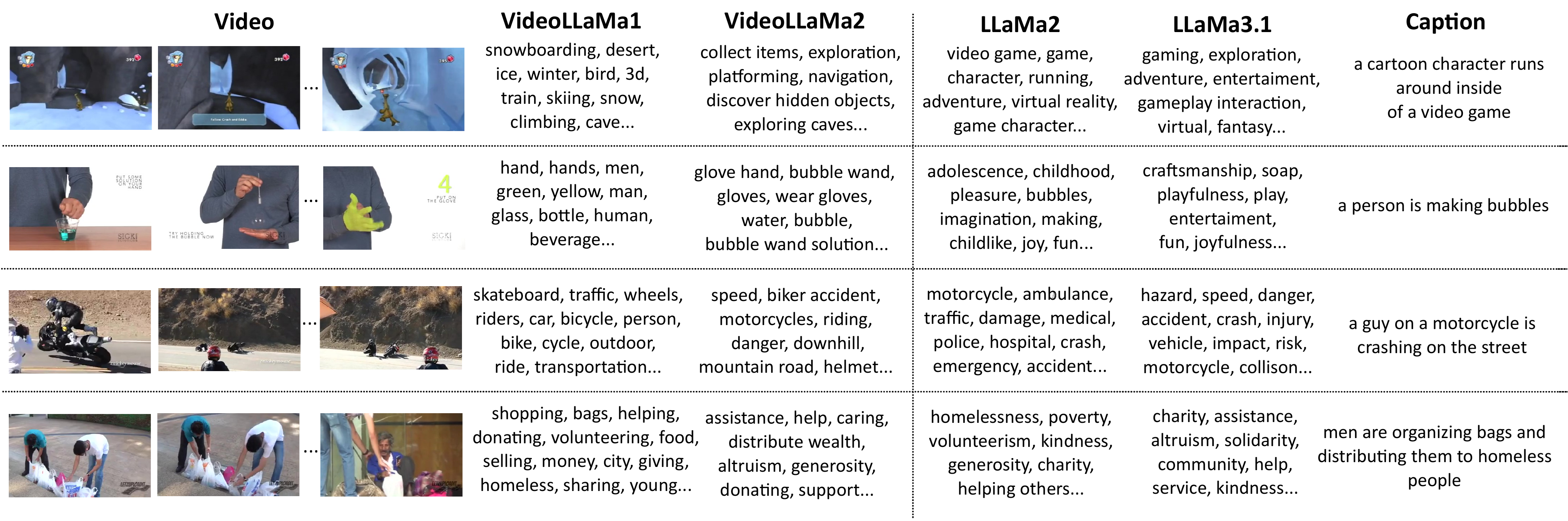}
\vspace*{-20pt}
    \caption{Comparison of extracted visual and textual tags on the MSR-VTT dataset when using different foundation models.}
    \label{fig:comaprison_qualitative_tags}
    \vspace*{-10pt}
\end{figure*}

As explained in Sec.~\ref{sec:ablation}, we considered different foundation models to extract visual and textual tags. More precisely, we considered Video-LLaMA~\cite{damonlpsg2023videollama} and  VideoLLaMA2~\cite{cheng2024videollama} to extract visual tags and Llama2~\cite{touvron2023llama2} and Llama3.1~\cite{DBLP:journals/corr/abs-2407-21783} to extract textual tags. We used the same parameters and same prompts to extract the tags by using all the considered foundations models.
Video-LLaMA and Llama2 tend to hallucinate tags more often as shown in Tab.~\ref{tab:comparison_stats_tags} and Fig.~\ref{fig:comaprison_qualitative_tags}.
More precisely, Tab.~\ref{tab:comparison_stats_tags} shows the statistics of extracted tags for MSR-VTT-9k. The total number of unique extracted visual tags by using Video-LLaMA~\cite{damonlpsg2023videollama} is much smaller than the total number obtained when using VideoLLaMA2~\cite{cheng2024videollama}. The same conclusion is reached for the total number of unique extracted textual tags by Llama2~\cite{touvron2023llama2} and Llama3.1~\cite{DBLP:journals/corr/abs-2407-21783} respectively. This shows the lesser ability of older foundation models to generate tags compared with more recent versions. 
In particular, this is more evident when we focused on the visual tags, where not only the total number of unique tags is smaller but also the average number of tags per pairs is the same, meaning that many tags are shared among pairs and so there are less unique tags able to distinguish all the video/caption pairs.

Fig.~\ref{fig:comaprison_qualitative_tags} shows a qualitative comparison of the extracted tags by using different foundation models.
It is evident that Video-LLaMA and Llama2 tend to hallucinate tags that are not relevant with what shown in the video and described in the caption. Moreover, the textual tags extracted by using Llama2 are very often words that already appear in the caption. 

\begin{table}[t!]
\centering
\begin{tabular}{lc|cc|cc}

&&\multicolumn{2}{c}{\#Unique Tags} 
& \multicolumn{2}{c}{Avg \#Tags per item}\\\cline{3-6}

\multicolumn{1}{l}{Foundation Model}
&\multicolumn{1}{c|}{Tags}
&\multicolumn{1}{c}{Train}
&\multicolumn{1}{c|}{Test}
&\multicolumn{1}{c}{Train}
&\multicolumn{1}{c}{Test}\\\hline

Video-LLaMA~\cite{damonlpsg2023videollama}& Visual&8,049&3,500&27,11&27,12\\
VideoLLaMA2~\cite{cheng2024videollama}& Visual &63,383&12,118&27.69&27.83\\\hline\hline
Llama2~\cite{touvron2023llama2}&Textual&162,571&5,058&14.96&15.20 \\
Llama3.1~\cite{DBLP:journals/corr/abs-2407-21783}&Textual&320,351&8,326&27.17&26.52 \\
\hline\hline
\end{tabular}
\caption{Comparison of statistics of tags when using different foundation models on MSR-VTT-9k.}
\label{tab:comparison_stats_tags}
\vspace{-10pt}
\end{table}
For example, given the captions \textit{a cartoon character runs around inside of a video game} and its corresponding video, we can see that Video-LLaMA and Llama2 hallucinate some visual tags such as \textit{snowboarding, desert, skiing, bird, climbing}---definitely irrelevant to what appear in the video---and textual tags such as \textit{video game, running, character, game character} that are already words that appear in the caption, therefore they do not add any additional information to better retrieve the correct video. On the contrary VideoLLaMA2 and Llama3.1 tend to extract tags that add additional information to the video and text.\\ See Fig.~\ref{fig:comaprison_qualitative_tags} for more examples on all the considered datasets.

\section{How to Generate Auxiliary Captions.}
\label{appendix:generate_new_captions}
In Sec.~\ref{sec:ablation}, we ablate the use of auxiliary captions instead of using tags.
We generate these additional captions by extracting them directly from the video and paraphrasing the original caption. We consider different approaches to extract captions from video and text.\\
\textbf{Visual Captions.} We consider two different approaches to generate new captions from a video: Blip2~\cite{li2023blip} and VideoLLaMA2~\cite{cheng2024videollama}.
Following the same approach proposed in~\cite{DBLP:conf/eccv/WangLLYHCPZWSJLXZHQWW24}, we generate new captions by extracting the middle frame of each video and use Blip2 to generate a new caption.
We use the same parameters of VideoLLaMA2 to extract captions as we did to extract visual tags in MAC-VR. 
However, we used a general prompt to ask VideoLLaMA2 to generate new captions. 
\begin{mdframed}[backgroundcolor=blue!10, linecolor=red, linewidth=2pt, roundcorner=10pt, shadow=true]

You are a conversational AI agent. You typically look at a video and generate a new caption for a video. Generate 10 new captions. Give me a bullet list as output.

\end{mdframed}
\vspace{5pt}
\textbf{Textual Captions.} We consider the paraphraser PEGASUS~\cite{DBLP:conf/icml/ZhangZSL20} and Llama3.1~\cite{DBLP:journals/corr/abs-2407-21783} to paraphrase the original caption. PEGASUS~\cite{DBLP:conf/icml/ZhangZSL20} is a standard Transformer-based encoder-decoder method pre-trained on a massive text corpora with a novel pre-training objective called Gap Sentence Generation (GSG). Instead of using traditional language modeling, PEGASUS removes important sentences from a document (gap-sentences) and asks the model to predict these missing sentences. After the pre-training stage, PEGASUS is fine-tuned on specific summarization datasets to improve its performance on downstream tasks. The model becomes highly effective at generating concise and accurate summaries by leveraging its pre-training knowledge.
\begin{figure*}[t!]
\centering
\includegraphics[width=\linewidth]{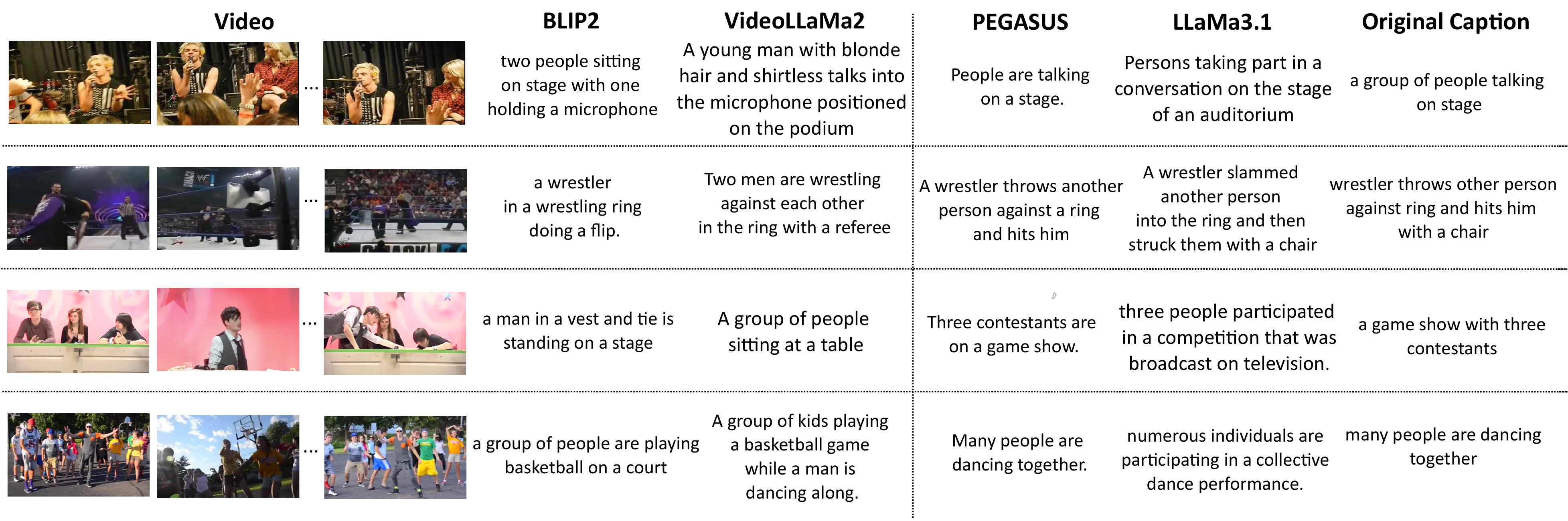}
    \vspace*{-20pt}
    \caption{Examples of extracted captions on the MSR-VTT dataset.}    \label{fig:auxiliary_captions_examples}
        \vspace*{-10pt}
\end{figure*}
We extract new captions from a caption by Llama3.1~\cite{DBLP:journals/corr/abs-2407-21783} by giving as input a general prompt as we did to extract tags: 
\begin{mdframed}[backgroundcolor=blue!10, linecolor=red, linewidth=2pt, roundcorner=10pt, shadow=true]

A chat between a curious user and an artificial intelligence assistant. The assistant gives helpful, detailed, and polite answers to the user's questions.
USER:
You are a conversational AI agent. You typically paraphrase sentences by using different words but keeping the same meaning.

Given the following sentence: 
1) \{\}

Generate 10 different sentences that are a paraphrased version of the original sentence. Give me a bullet list as output.

ASSISTANT:

\end{mdframed}
\vspace{5pt}
We randomly pick an extracted visual and textual caption as auxiliary inputs in MAC-VR during training, noting a similar length of input to our `sentence of tags'. In inference, we always pick the first caption in the set of the extracted ones. Some examples of the extracted captions with all the considered methods are shown in Fig.~\ref{fig:auxiliary_captions_examples}.

\section{Additional t-SNE plot.}
\label{appendix:additional_tsne}
\begin{figure*}[ht!]
    \begin{minipage}[b]{0.49\textwidth}
        \centering
        \includegraphics[width=\linewidth]{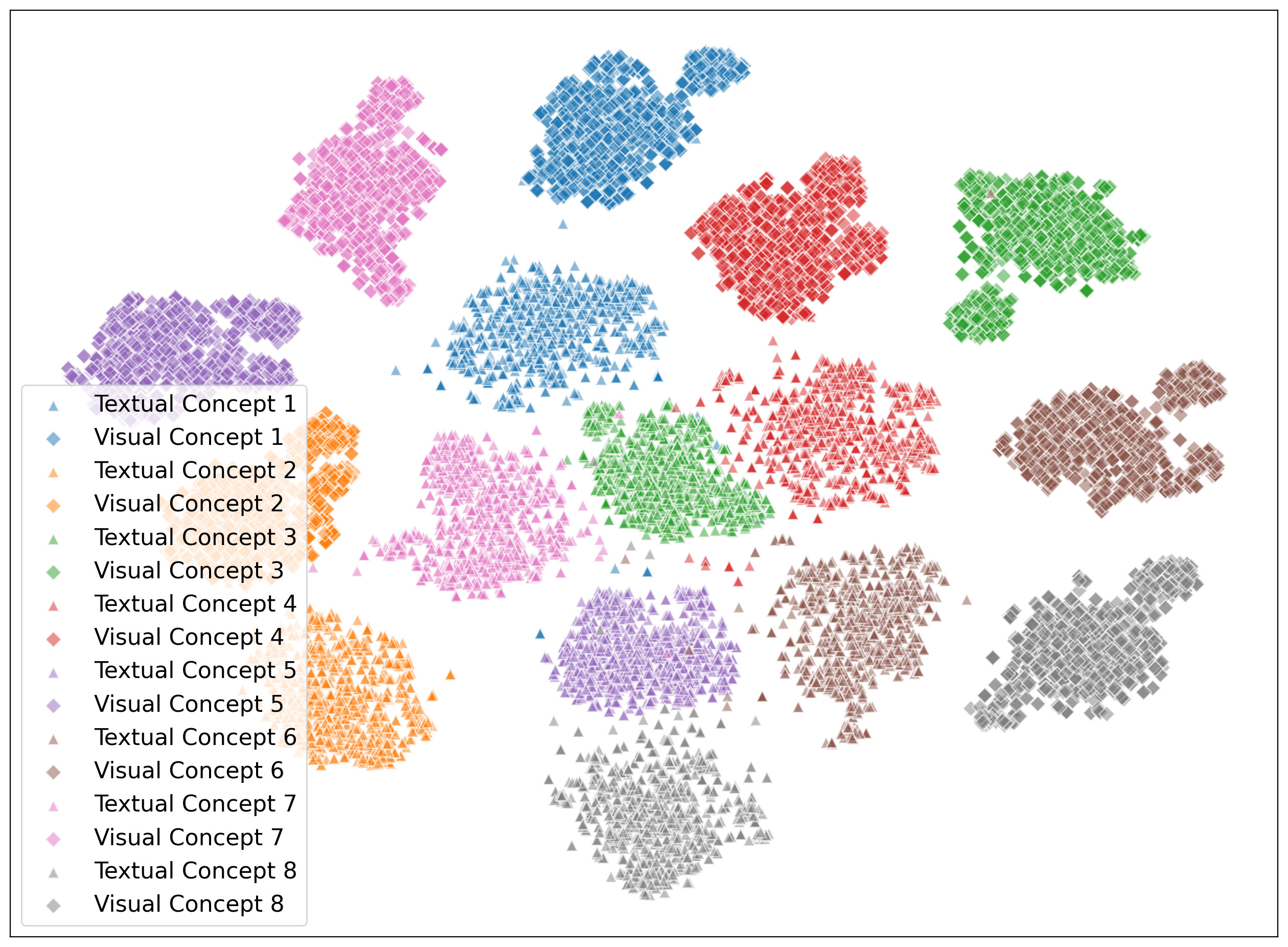}
    \end{minipage}
    \begin{minipage}[b]{0.49\textwidth}
        \centering
        \includegraphics[width=\linewidth]{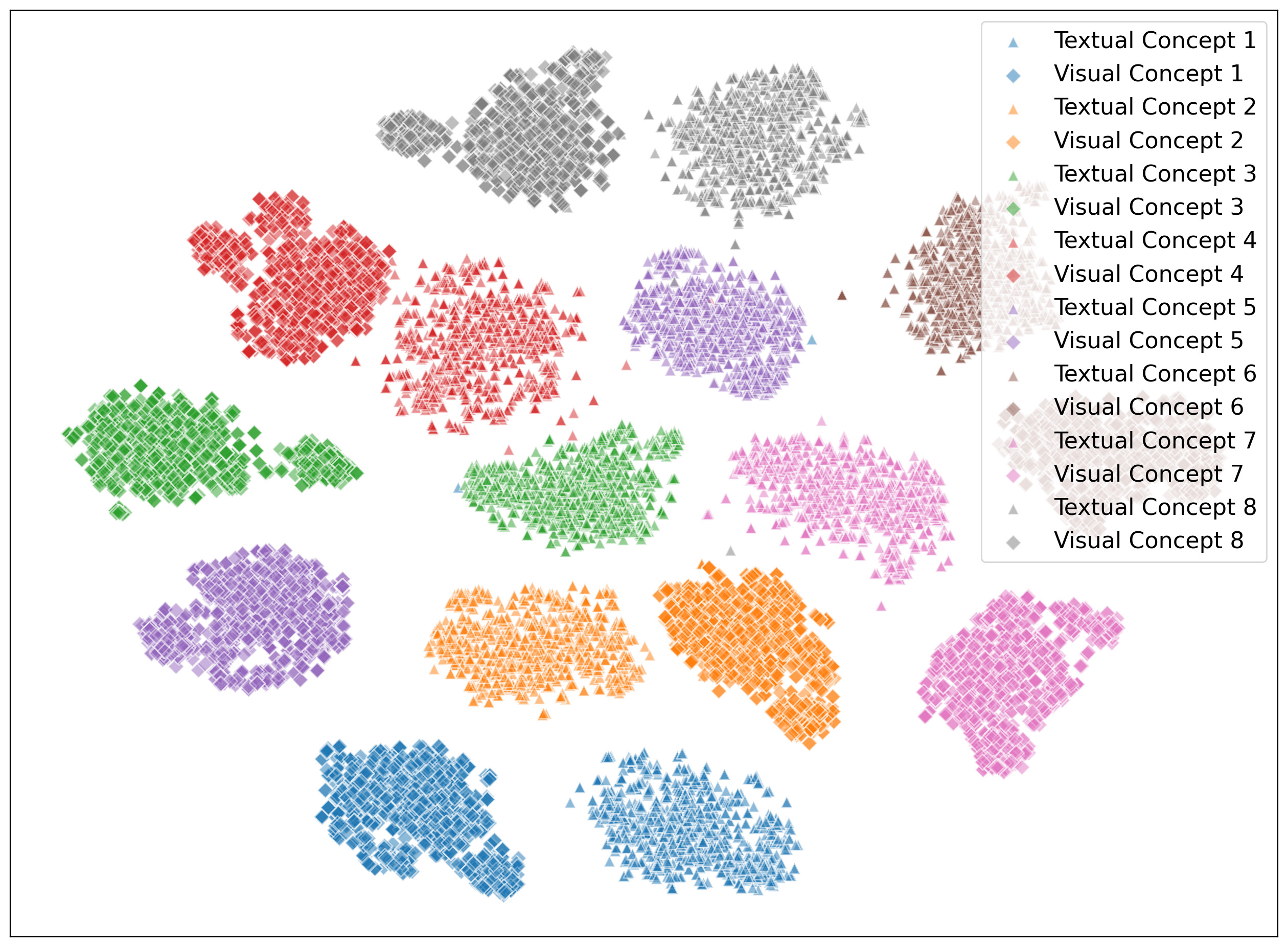}
    \end{minipage}
\caption{t-SNE plot of visual and textual concepts on  DiDeMo without (\textbf{left}) and with using auxiliary modality-specific tags (\textbf{right}).}
\label{fig:tsne_plot_didemo}
\end{figure*}

\begin{figure*}[ht!]
\centering
\begin{minipage}[b]{0.49\textwidth}
    \centering
    \includegraphics[width=\linewidth]{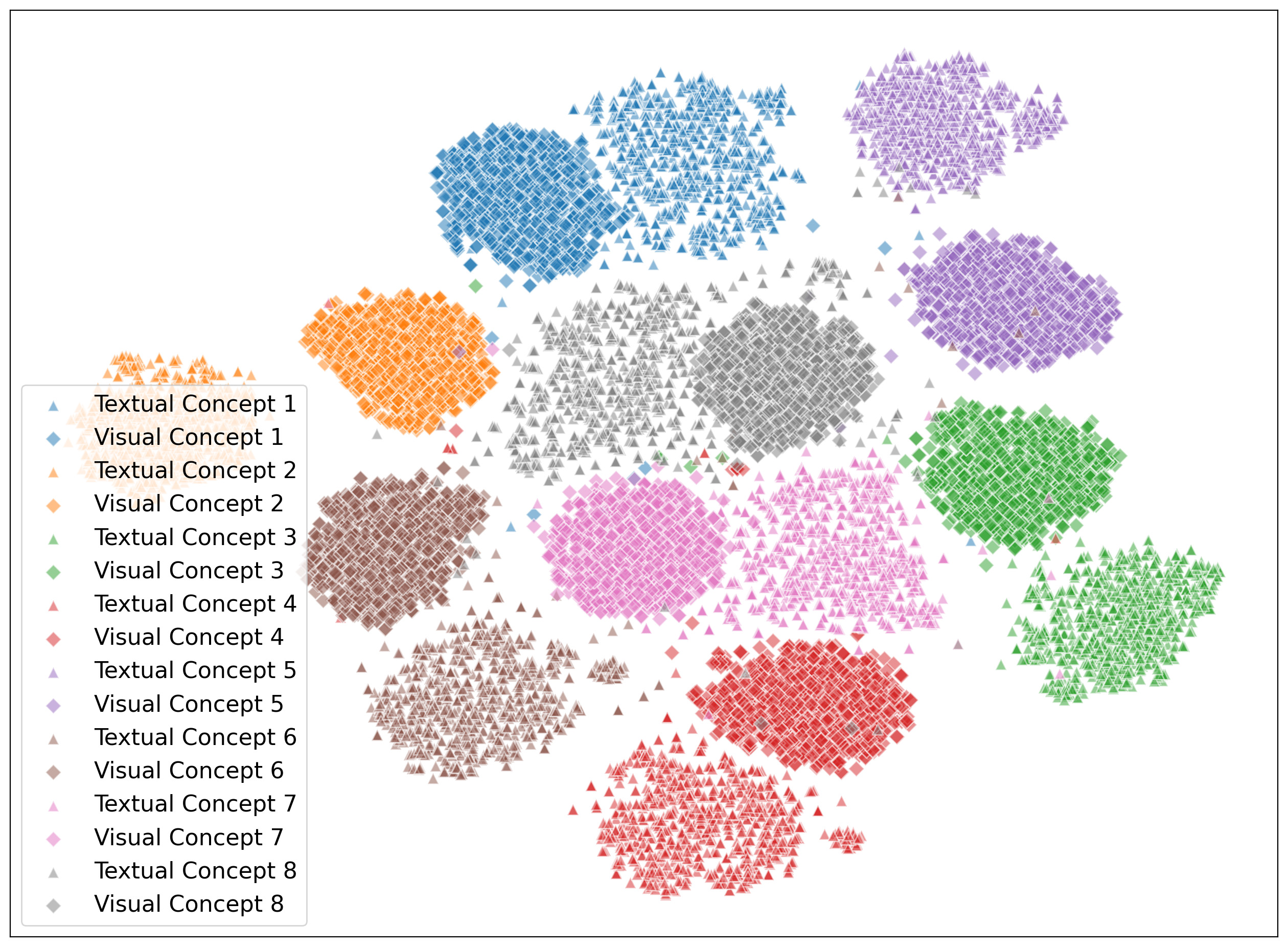}
\end{minipage}
\hfill
\begin{minipage}[b]{0.49\textwidth}
    \centering
    \includegraphics[width=\linewidth]{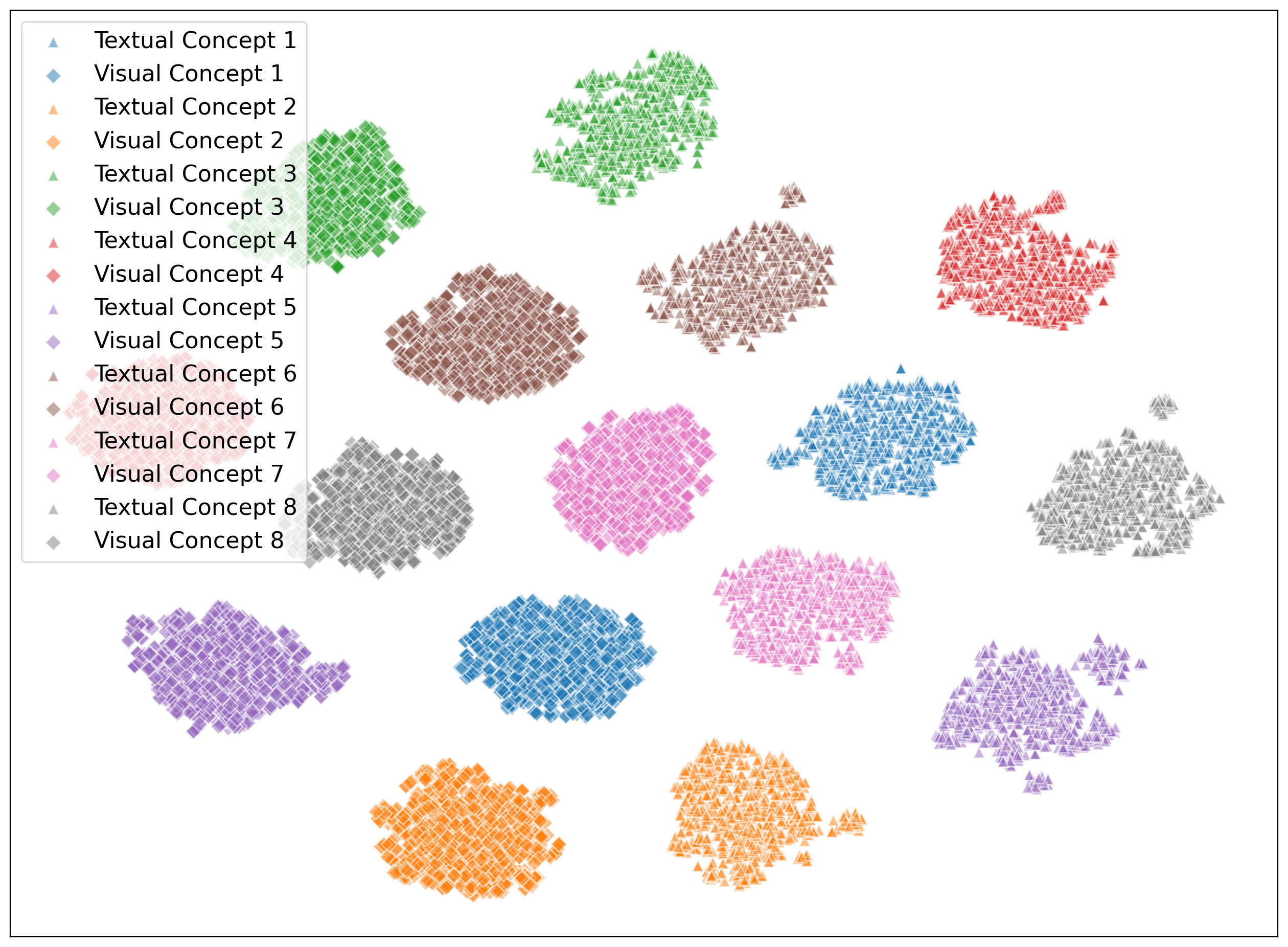}
\end{minipage}
\caption{t-SNE plot of visual and textual concepts on  Charades without (\textbf{left}) and with using auxiliary modality-specific tags (\textbf{right}).}
\label{fig:tsne_plots_charades}
\end{figure*}

\begin{figure*}[ht!]
    \begin{minipage}[b]{0.49\textwidth}
        \centering
        \includegraphics[width=\linewidth]{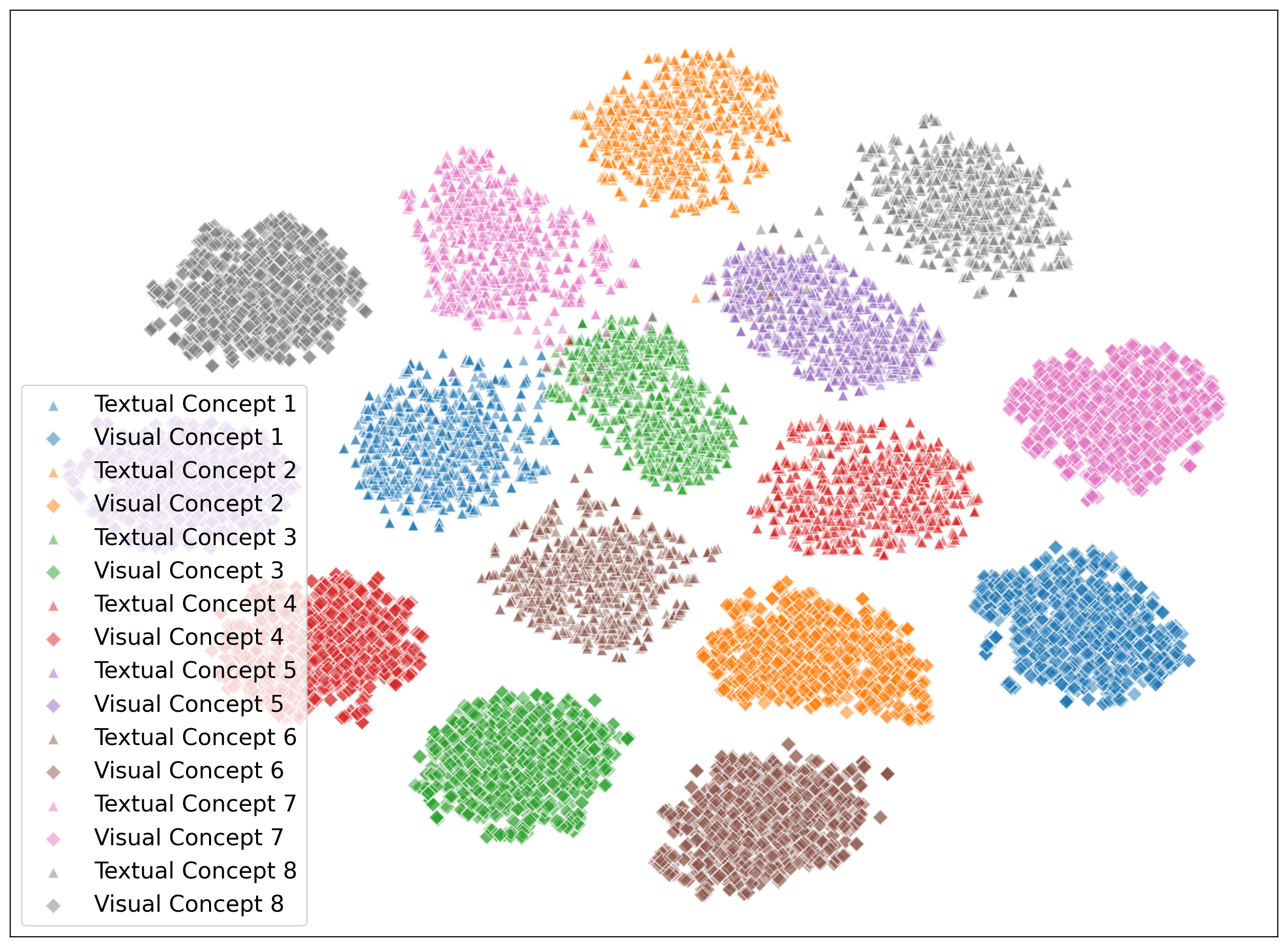}
    \end{minipage}
    \begin{minipage}[b]{0.49\textwidth}
        \centering
        \includegraphics[width=\linewidth]{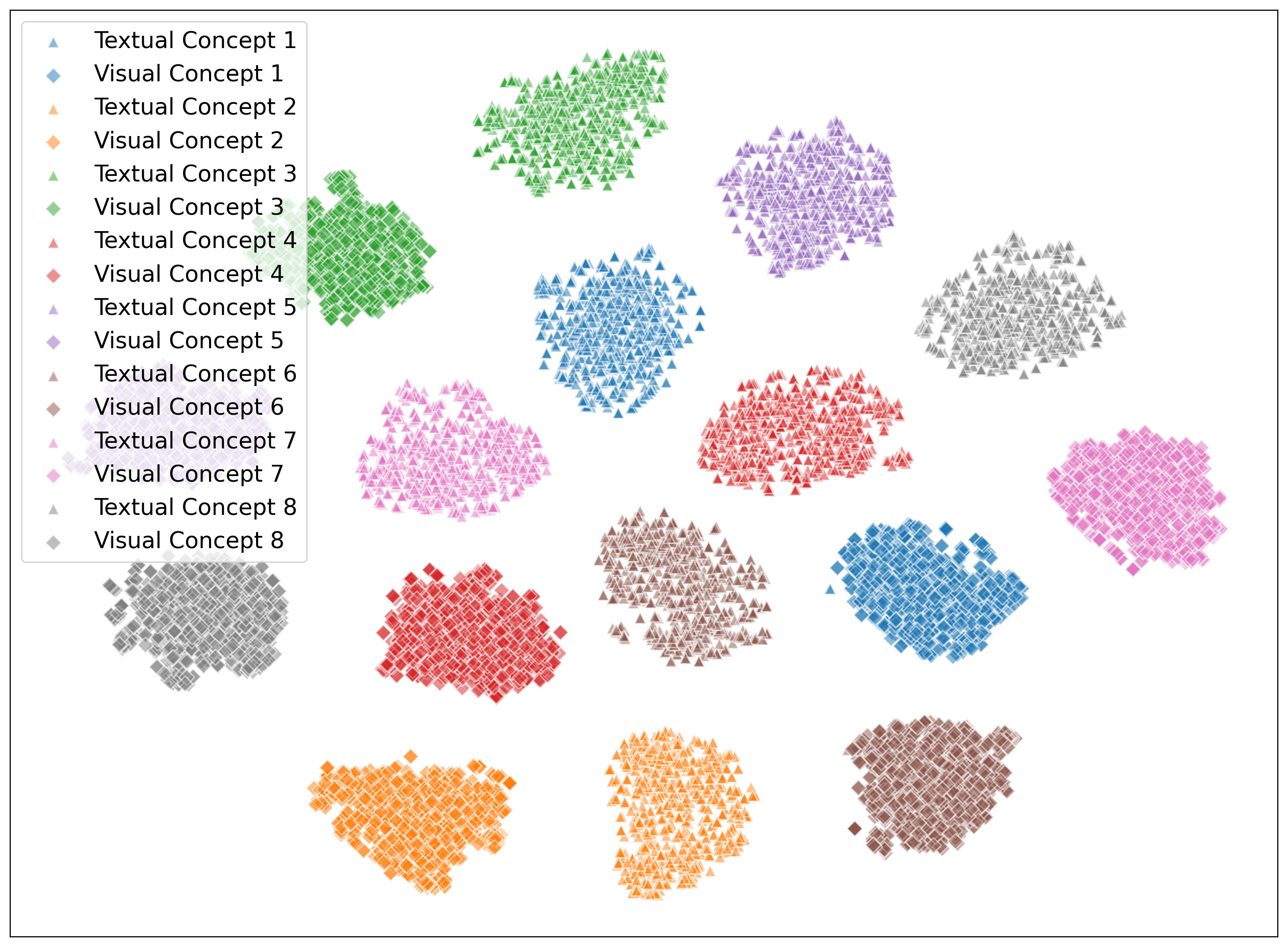}
    \end{minipage}
\caption{t-SNE plot of visual and textual concepts on  TGIF without (\textbf{left}) and with using auxiliary modality-specific tags (\textbf{right}).}
\label{fig:tsne_plot_tgif}
\end{figure*}

\begin{figure*}[ht!]
    \begin{minipage}[b]{0.49\textwidth}
        \centering
        \includegraphics[width=\linewidth]{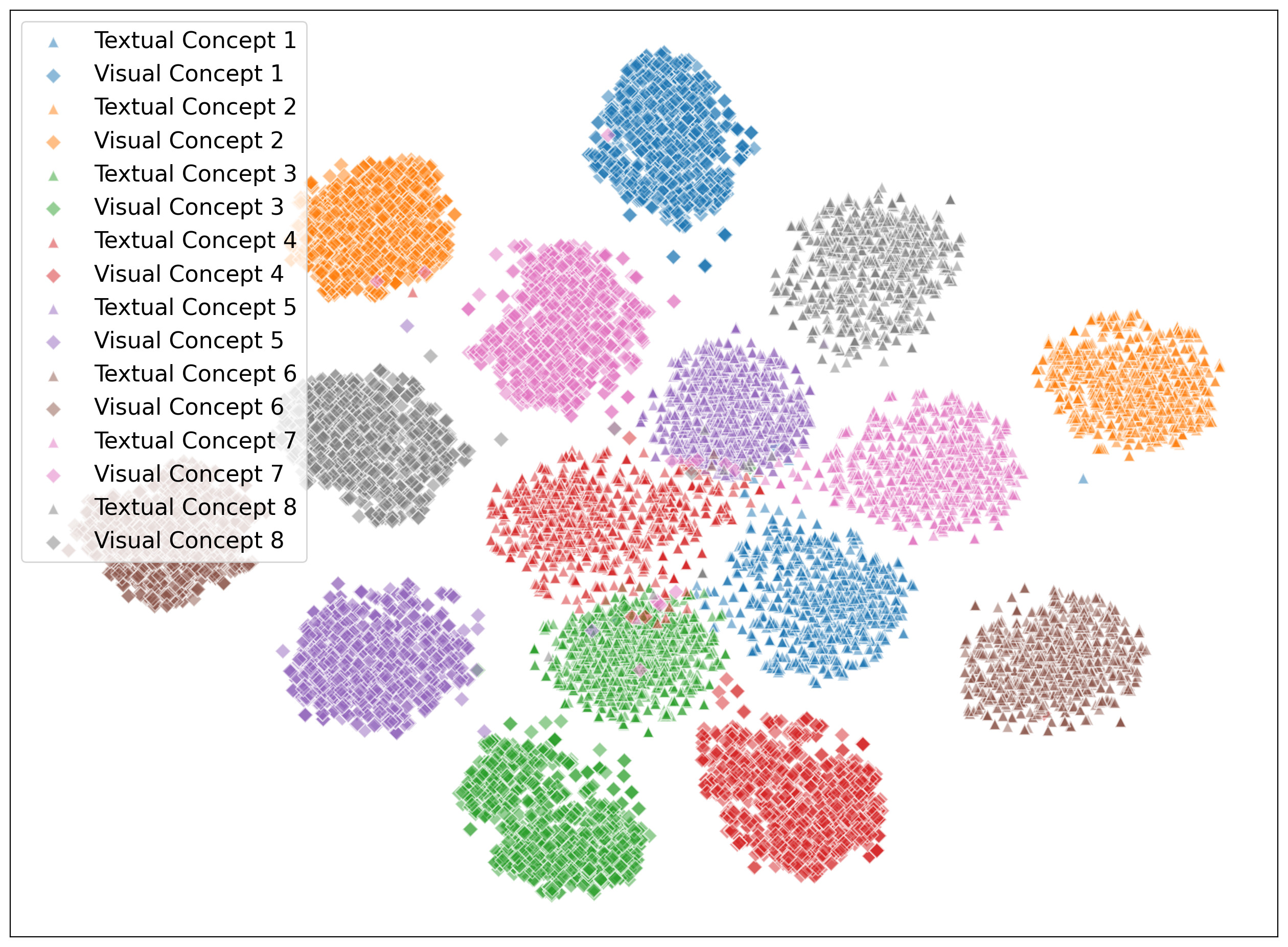}
    \end{minipage}
    \begin{minipage}[b]{0.49\textwidth}
        \centering
        \includegraphics[width=\linewidth]{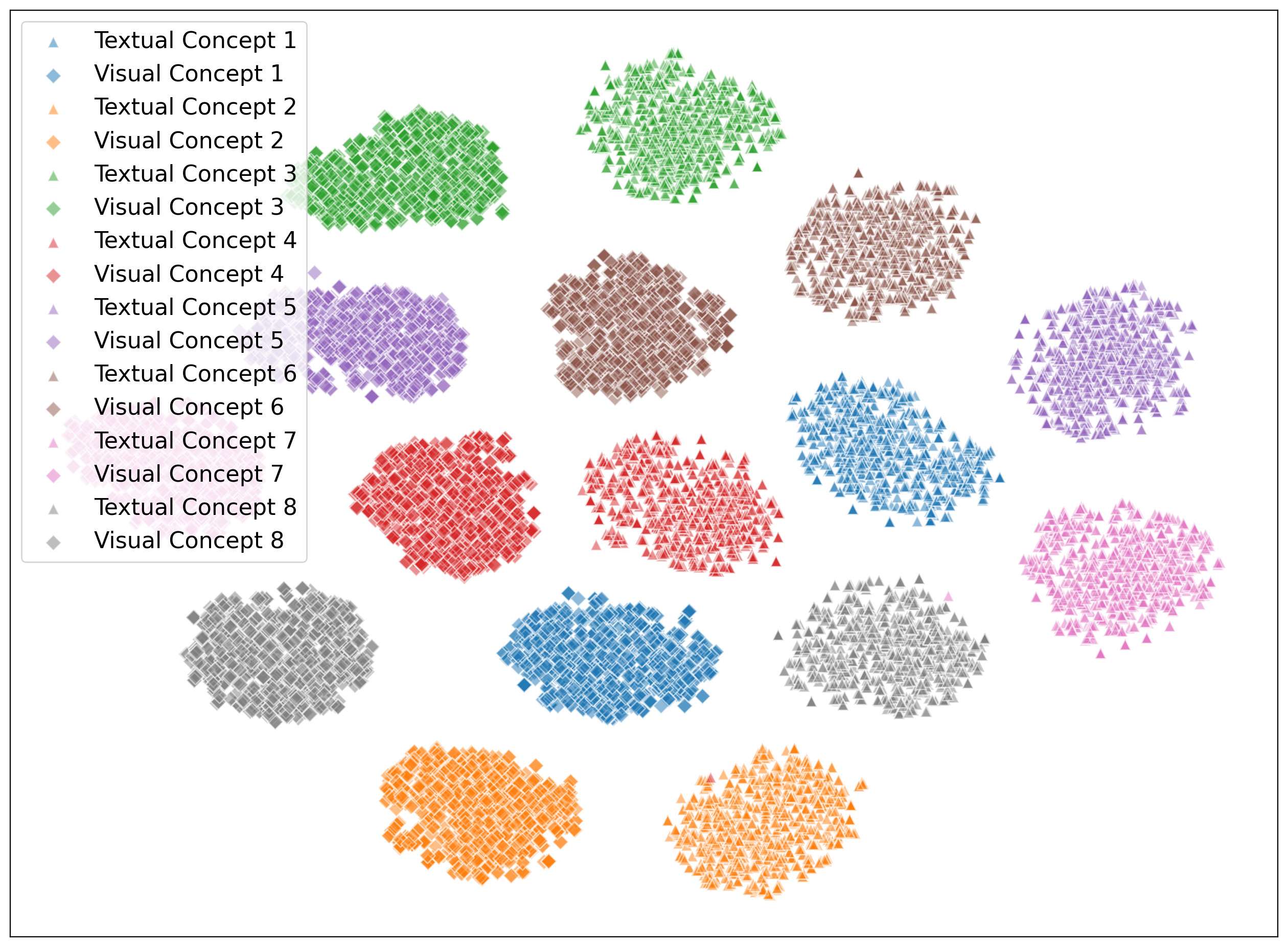}
    \end{minipage}
\caption{t-SNE plot of visual and textual concepts on  YouCook2 without (\textbf{left}) and with using auxiliary modality-specific tags (\textbf{right}).}
\label{fig:tsne_plot_yc2}
\end{figure*}

Fig.~\ref{fig:tsne_plot_didemo} to~\ref{fig:tsne_plot_yc2} show the t-SNE plot of MAC-VR of visual and textual concepts with/without auxiliary modality-specific tags across all the other datasets.  We sample $1,000$ pairs for Charades, TGIF and YouCook2. We can see that, also for these datasets, the use of auxiliary modality-specific tags helps to better distinguish the different concepts and in the t-SNE plot. 

\begin{figure*}[ht!]
        \centering
    \begin{minipage}[b]{0.49\textwidth}
        \centering
        \includegraphics[width=\linewidth]{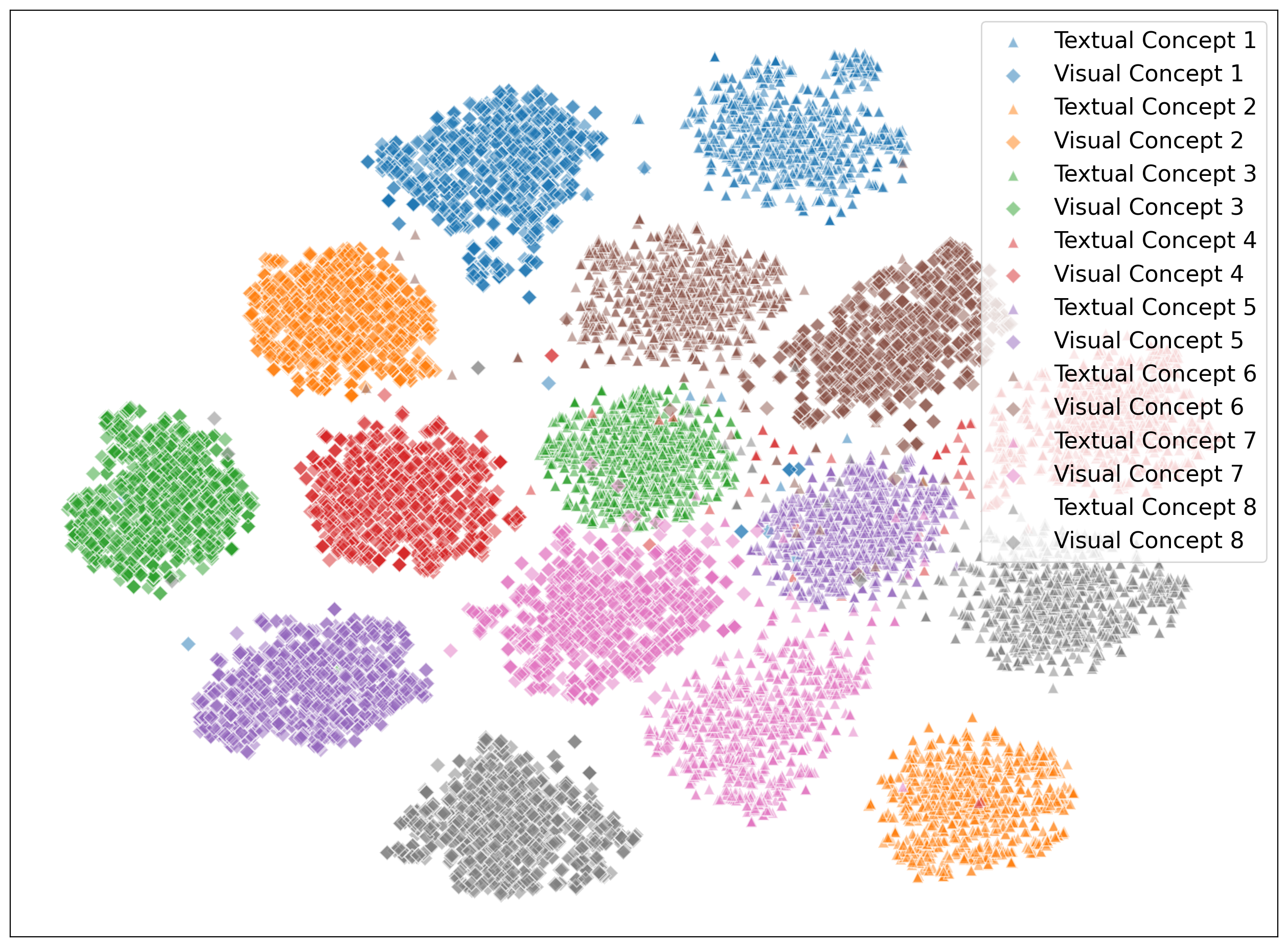}
    \end{minipage}
    \begin{minipage}[b]{0.49\textwidth}
        \centering
        \includegraphics[width=\linewidth]{images/MAC-VR-MSRVTT-concepts.jpg}
    \end{minipage}
\caption{t-SNE plot of visual and textual concepts on  MSR-VTT-9k without Alignment Loss $\mathcal{L}_{A}$ (\textbf{left}) and with Alignment Loss $\mathcal{L}_{A}$ (\textbf{right})..}
\label{fig:tsne_alignment_loss}
\end{figure*}

\begin{figure}[ht!]
    \begin{minipage}[b]{0.49\textwidth}
        \centering
        \includegraphics[width=\linewidth]{images/DiCoSA-MSRVTT-concepts.jpg}
    \end{minipage}
    \begin{minipage}[b]{0.49\textwidth}
        \centering
        \includegraphics[width=\linewidth]{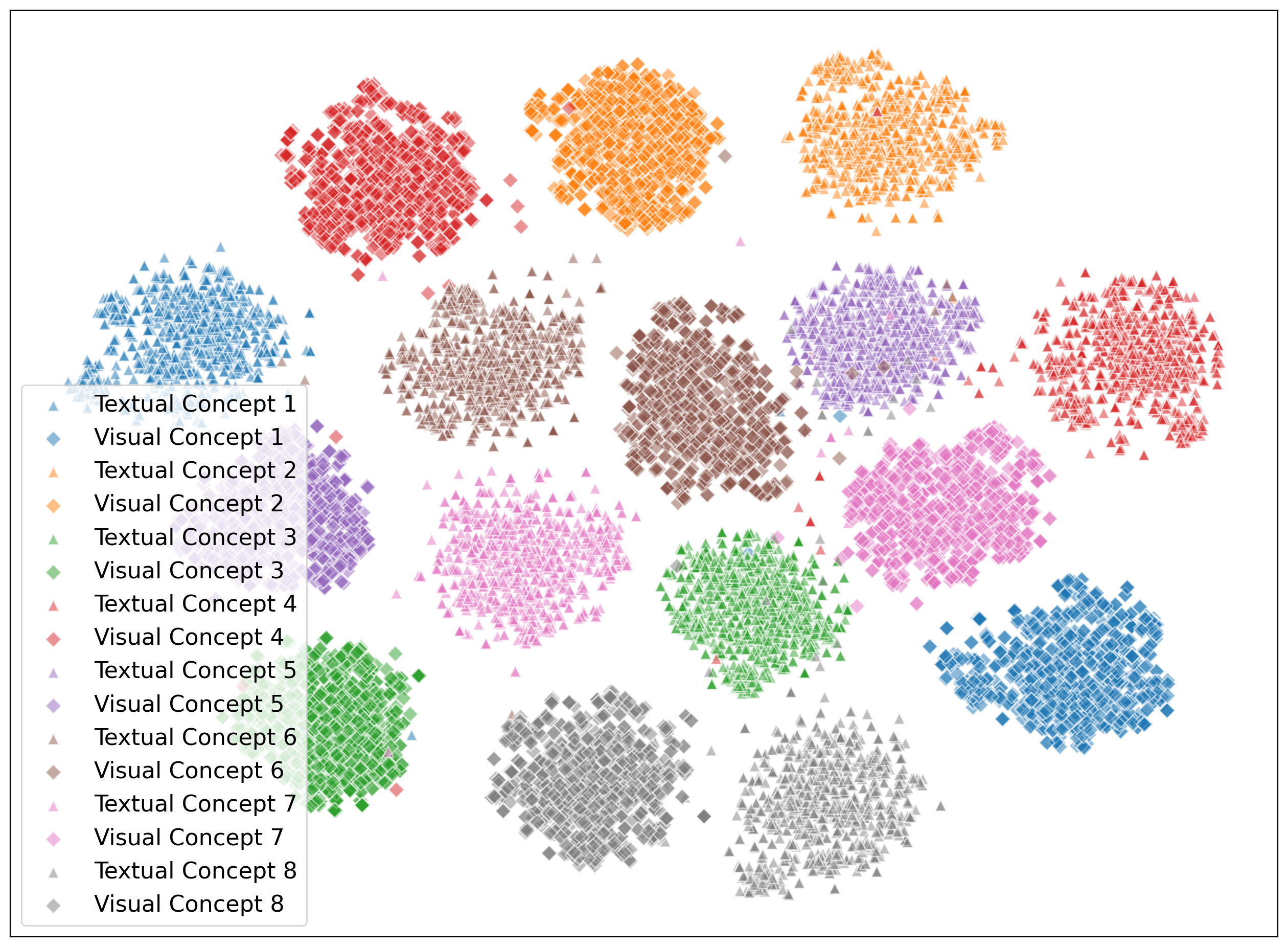}
    \end{minipage}
\caption{t-SNE plot of visual and textual concepts on  MSR-VTT-9k without auxiliary modality-specific tags (\textbf{left}) and with using only visual tags  (\textbf{right}).}
\label{fig:tsne_only_visual}
\vspace*{-10pt}
\end{figure}

\begin{figure*}[ht!]
    \begin{minipage}[b]{0.49\textwidth}
        \centering
        \includegraphics[width=\linewidth]{images/DiCoSA-MSRVTT-concepts.jpg}
    \end{minipage}
    \begin{minipage}[b]{0.49\textwidth}
        \centering
        \includegraphics[width=\linewidth]{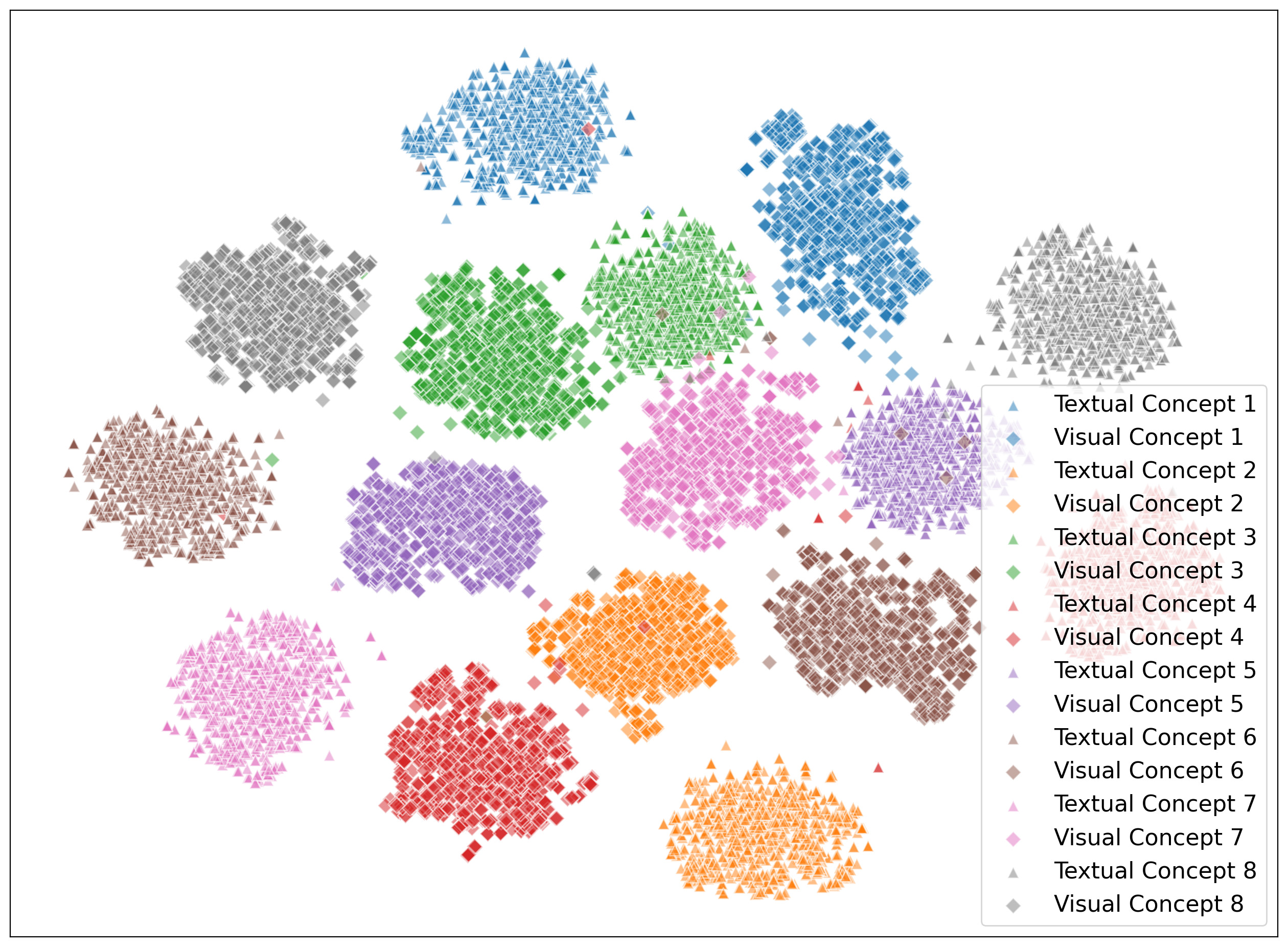}
    \end{minipage}
\caption{t-SNE plot of visual and textual concepts on  MSR-VTT-9k without auxiliary modality-specific tags (\textbf{left}) and with using only textual tags  (\textbf{right}).}
\label{fig:tsne_only_textual}
\vspace*{-10pt}
\end{figure*}

In Fig.~\ref{fig:tsne_alignment_loss}, we show the t-SNE plot of MAC-VR on the MSR-VTT-9k test set when using modality-specific tags with/without our Alignment Loss $\mathcal{L}_{A}$. It is clear that the use of $\mathcal{L}_{A}$ helps to better distinguish the different concepts and have better clusters in the t-SNE plot.
In Fig.~\ref{fig:tsne_only_visual} and Fig.~\ref{fig:tsne_only_textual}, we show the t-SNE plot of visual and textual concepts without auxiliary modality-specific tags and when using only visual tags (Fig.~\ref{fig:tsne_only_visual}) and textual tags (Fig.~\ref{fig:tsne_only_textual}) with our Alignment Loss $\mathcal{L}_{A}$. From this, both tags used individually help to better align the visual and textual concepts.
In particular, the visual tags help to better align the visual and textual concepts compared to the textual tags. A possible explanation for this is that the tags extracted from videos share the same modality as captions, which facilitates better alignment between visual and textual concepts. We leave this conclusion as possible inspiration for future works in this field.

\section{Qualitative Results}
\label{appendix:additional_qualitative}

\begin{figure*}[h!]
\centering
\includegraphics[width=\linewidth]{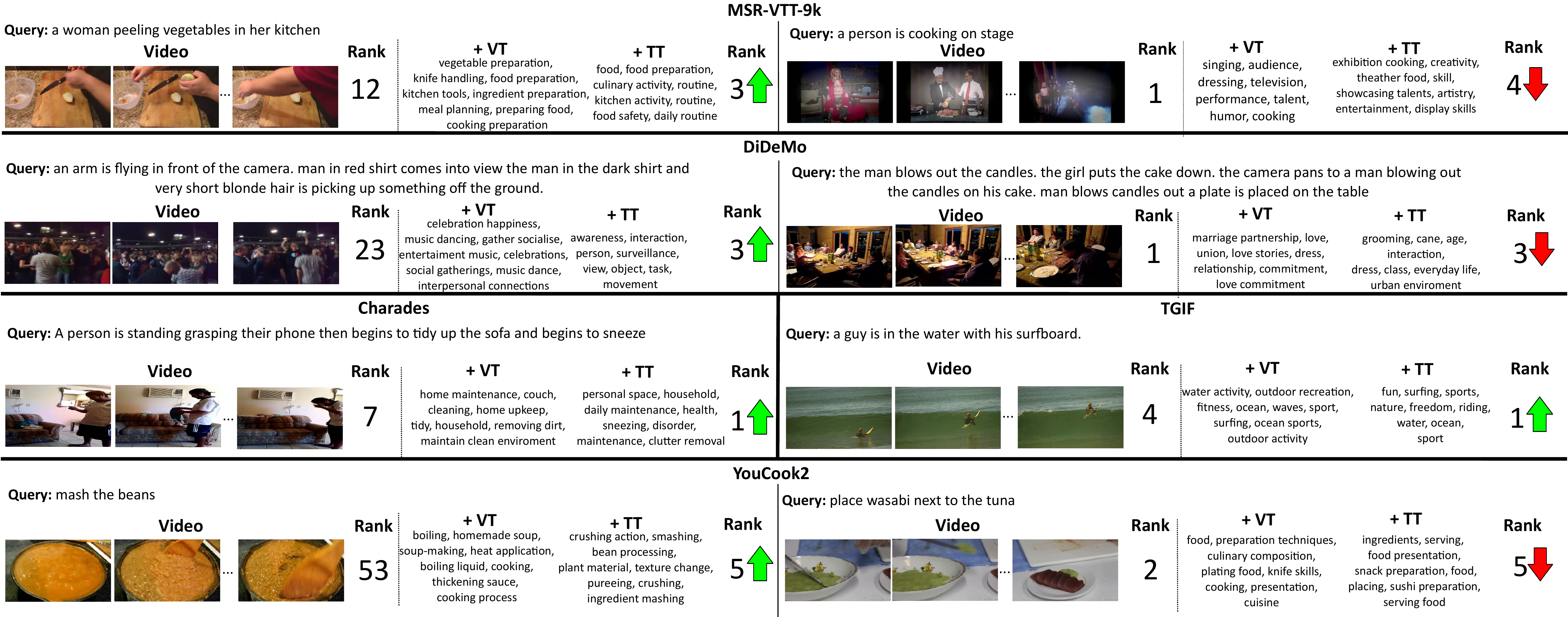}
        \vspace*{-10pt}
    \caption{Additional qualitative results across datasets.}
    \label{fig:sup_qualitative_results}
\end{figure*}

In Fig.~\ref{fig:sup_qualitative_results}, we show additional qualitative results across all the datasets. It is evident that tags add additional information that is not described in the text. For example, the video associated to the caption \textit{a guy is in the water with his surfboard} has visual (i.e. \textit{water activity, ocean, waves, surfing, outdoor activity}) and textual (i.e. \textit{surfing, water, ocean, riding}) tags that add additional information and boost the rank position by $3$. Similarly, in the YouCook2 example, the video is associated to a general \textit{mash the beans}, tags help to add additional information, indeed visual tags such as \textit{boiling, homemade soup, boiling liquid, thickening sauce} and textual tags such as such as \textit{crushing action, bean processing, 
pureeing, ingredient mashing} add semantic information that definitely help the model in the retrieval process.
Rarely, the extracted tags were not semantically relevant to what was shown in the video or described in the caption. For instance, the caption \textit{the man blows out the candles. the girl puts the cake down. the camera pans to a man blowing out the candles on his cake. man blows candles out a plate is placed on the table} has visual tags such as \textit{marriage partnership, love, love commitment} and textual tags such as \textit{class}. Another example is the query \textit{a person is cooking on stage}, where we can see that \textit{singing} is one of the visual tags extracted. The VLM extracts this tag from the last frame where there is a person singing on stage as the scene changes.
In other cases, visual and textual tags are too general, even though they are relevant to the video and caption, and so they do not add any additional semantic information. For example, the video associated with the caption \textit{place wasabi next to the tuna} has \textit{food, cooking, preparation techniques} as visual tags and \textit{ingredients, serving, serving food} as textual tags. Even though these tags are relevant to the video and caption, they are too general and do not add any additional semantic knowledge.

\end{document}